\RequirePackage{fix-cm}

\documentclass[smallcondensed]{svjour3}

\smartqed  

\usepackage{graphicx}
\usepackage{url,hyperref,lineno,microtype,subcaption}
\usepackage{amsmath,mathtools,amsfonts,amssymb}
\usepackage{algorithm}
\usepackage[noend]{algpseudocode}
\usepackage[onehalfspacing]{setspace}
\usepackage{comment}
\usepackage{natbib}

\floatname{algorithm}{Algorithm}

\algnewcommand{\IfSingle}[1]{\State\algorithmicif\ #1\ \algorithmicthen}
\algnewcommand{\ElseSingle}[1]{\State\algorithmicelse\ #1}

\newenvironment{mysplit}%
  {\arraycolsep 0pt \begin{array}{l}}%
  {\end{array}}

\def\folhappensAt{\mathtt{HappensAt}}
\def\folholdsAt{\mathtt{HoldsAt}}
\def\folinitiatedAt{\mathtt{InitiatedAt}}
\def\folterminatedAt{\mathtt{TerminatedAt}}

\def\move{$\mathtt{move}$}
\def\meet{$\mathtt{meet}$}
\def\rendezvous{$\mathtt{rendezvous}$}

\def\PEC{\mathtt{MLN}{\!-\!}\mathtt{EC}}
\def\MLNEC{$\PEC$}
\def\OSLa{OSL$\alpha$}

\begin{document}

\title{Semi-Supervised Online Structure Learning for Composite Event Recognition}


\author{Evangelos Michelioudakis \and 
        Alexander Artikis \and 
        Georgios Paliouras}

\institute{Evangelos Michelioudakis \at
           Inst. of Informatics \& Telecom., 
           National Center for Scientific Research ``Demokritos'' and \\
           Dept. of Informatics \& Telecom., 
           National and Kapodistrian University of Athens \\
           \email{vagmcs@iit.demokritos.gr}
           \and
           Alexander Artikis \at
           Dept. of Maritime Studies, University of Piraeus and \\
           Inst. of Informatics \& Telecom., 
           National Center for Scientific Research ``Demokritos'' \\
           \email{a.artikis@iit.demokritos.gr}
           \and
           Georgios Paliouras \at
           Inst. of Informatics \& Telecom., 
           National Center for Scientific Research ``Demokritos'' \\
           \email{paliourg@iit.demokritos.gr}
}

\date{Received: date / Accepted: date}

\maketitle

\begin{abstract}
Online structure learning approaches, such as those stemming from Statistical Relational Learning, enable the discovery of complex relations in noisy data streams. However, these methods assume the existence of fully-labelled training data, which is unrealistic for most real-world applications. We present a novel approach for completing the supervision of a semi-supervised structure learning task. We incorporate graph-cut minimisation, a technique that derives labels for unlabelled data, based on their distance to their labelled counterparts. In order to adapt graph-cut minimisation to first order logic, we employ a suitable structural distance for measuring the distance between sets of logical atoms. The labelling process is achieved online (single-pass) by means of a caching mechanism and the Hoeffding bound, a statistical tool to approximate globally-optimal decisions from locally-optimal ones. We evaluate our approach on the task of composite event recognition by using a benchmark dataset for human activity recognition, as well as a real dataset for maritime monitoring. The evaluation suggests that our approach can effectively complete the missing labels and eventually, improve the accuracy of the underlying structure learning system.

\end{abstract}

\section{Introduction} \label{sec:intro}

Methods for handling both uncertainty and complex relational structure have received much attention in machine learning. For instance, in composite event recognition \citep{CugolaM12,ArtikisSPP12,AlevizosSAP17}, relations are defined over entities of actors and objects involved in an event. Such applications are typically characterised by uncertainty, and in many cases data of significant volume and velocity. Manual derivation of relational dependencies is a time-consuming process and, in the presence of large data streams, unrealistic.

One of the logic-based representations that handles uncertainty is Markov Logic Networks (MLNs) \citep{domingos2006markov} that combine first-order logic and probabilistic graphical models. Online structure learning approaches for MLNs have been effectively applied to a variety of tasks \citep{vagmcs2016osla,huynh2011osl}. Although these approaches facilitate the automated discovery of multi-relational dependencies in noisy environments, they assume a fully labelled training sequence, which is unrealistic in most real-world applications.

We propose a novel method for completing the supervision, using the graph-cut minimisation technique \citep{ZhuSemiSupervised} and a distance function for first-order logic. Graph-cut minimisation essentially derives labels for unlabelled data, by computing their distance to their labelled counterparts. In particular, we adapt the graph-cut minimisation approach proposed by \cite{ZhuGL03} to first-order logic, in order to operate over logical structures instead of numerical data. To do so, we use a structural measure \citep{NienhuysCheng}, designed to compute the distance between logical atoms, and modify it using the Kuhn-Munkres algorithm \citep{Kuhn1955}, to accurately calculate the distance over sets of logical atoms that represent the training examples.

The proposed supervision completion method operates in an online fashion (single-pass), by means of a caching mechanism that stores previously seen labels for future usage. The Hoeffding bound, a statistical tool that enables approximate globally-optimal decisions from locally-optimal ones, is used to filter out contradicting labels that may compromise the labelling accuracy. The completed training data can be subsequently used by any supervised structure learner. To demonstrate the benefits of SPLICE, our proposed method to semi-supervised online structure learning, we focus on \emph{composite event recognition} (CER), by employing the \OSLa\ \citep{vagmcs2016osla} and OLED \citep{KatzourisAP16} online structure learners. Both of these learners can construct Event Calculus theories \citep{KowalskiS86,mueller2008,anskarlTOCL15} for CER applications.

In CER, the goal is to recognise \textit{composite events} (CEs) of interest, given an input stream of \textit{simple derived events} (SDEs). CEs can be defined as relational structures over sub-events, either CEs or SDEs, and capture the knowledge of a target application. The proposed method (SPLICE) is evaluated on the task of activity recognition from surveillance video footage, as well as in maritime monitoring. In the former case, the goal is to recognise activities taking place between persons, e.g., people meeting or moving together, by exploiting information about observed activities of individuals. In maritime monitoring, the goal is to recognise vessel activities, by exploiting information such as vessel speed, location and communication gaps. Our empirical analysis suggests that our approach is capable of completing the supervision, even in the presence of little given annotation, and eventually enhances the accuracy of the underlying structure learner.

In summary, the main contributions of this paper are:

\begin{enumerate}
    \item An online supervision completion method using a caching mechanism to store labelled examples for future usage, and the Hoeffding bound to filter out contradicting examples that may compromise the overall accuracy.
    
    \item An adaptation of the graph-cut minimisation technique to first-order logic, using a structural distance for comparing logical atoms, and the Kuhn-Munkres algorithm for improving the accuracy of the distance calculation.
    
    \item The first system for semi-supervised online structure learning combining online supervision completion and two state-of-the-art structure learners, in order to learn Event Calculus definitions for CER.

    \item An evaluation of the combined system on two real (non-synthetic) datasets concerning activity recognition and maritime monitoring. In the former case, we use a benchmark video surveillance dataset that includes manually constructed ground truth. In the latter case, we use a dataset comprising vessel position signals from the area of Brest, France, spanning one month.
\end{enumerate}

The remainder of the paper is organised as follows. Section \ref{sec:background} provides the required background for the proposed method and Section \ref{sec:ssl} describes our approach to semi-supervised online structure learning. Section \ref{sec:evaluation} reports the experimental results on both datasets. Section \ref{sec:related_work} discusses related work on semi-supervised structure learning and alternative distance measures for logical representations, while Section \ref{sec:conclusions} concludes and proposes directions for future research.

\section{Background} \label{sec:background}

We present existing methods that are employed in the rest of the paper. We begin by briefly presenting the Event Calculus, as well as the basic functionality of \OSLa\ and OLED structure learners. Then, in Section \ref{sec:graph.cut} we describe the ideas behind graph-cut minimisation and a variation based on the harmonic function. Finally, in Section \ref{sec:metric} we discuss a distance function used for comparing sets of logical atoms and one of its drawbacks that we overcome using the Kuhn-Munkres algorithm.

\subsection{Event Calculus and Structure Learning} \label{sec:ec_learning}

One way of performing CER is by using the \textit{discrete} Event Calculus (DEC) \citep{mueller2008}. The ontology of DEC consists of \textit{time-points}, \textit{events} and \textit{fluents}. The underlying time model is linear and represented by integers. A \textit{fluent} is a property whose value may change over time by the occurrence of particular \textit{events}. DEC includes the core domain-independent axioms of the Event Calculus, which determine whether a fluent holds or not at a specific time-point. This axiomatisation incorporates the common sense \emph{law of inertia}, according to which fluents persist over time, unless they are affected by an event occurrence. Event occurrences are denoted by the $\folhappensAt$ predicates, while $\folholdsAt$ predicates denote whether a fluent holds. The $\folinitiatedAt$ and $\folterminatedAt$ predicates express the conditions in which a fluent is initiated or terminated, and are triggered by $\folhappensAt$ predicates. The core DEC axioms are defined as follows:

\begin{minipage}{.45\linewidth}
  \begin{align}
  & \begin{mysplit}
  \label{axiom:dec_holdsAt}
  \mathtt{HoldsAt}(f,t{+}1) \Leftarrow\\
    \qquad \mathtt{InitiatedAt}(f,t)
  \end{mysplit} \\
  & \begin{mysplit}
  \label{axiom:dec_holdsAt_inertia}
  \mathtt{HoldsAt}(f,t{+}1) \Leftarrow\\
    \qquad \mathtt{HoldsAt}(f,t)\ \wedge \\
    \qquad \neg \mathtt{TerminatedAt}(f,t)
  \end{mysplit}
  \end{align}
\end{minipage}%
\begin{minipage}{.45\linewidth}
  \begin{align}
  & \begin{mysplit}
  \label{axiom:dec_not_holdsAt}
  \neg \mathtt{HoldsAt}(f,t{+}1) \Leftarrow\\
  \qquad \mathtt{TerminatedAt}(f,t)
  \end{mysplit} \\
  & \begin{mysplit}
  \label{axiom:dec_not_holdsAt_inertia}
  \neg \mathtt{HoldsAt}(f,t{+}1) \Leftarrow\ \\
  \qquad \neg \mathtt{HoldsAt}(f,t)\ \wedge \\
  \qquad \neg \mathtt{InitiatedAt}(f,t)
  \end{mysplit}
  \end{align}
\end{minipage}\\

\noindent Variables and functions start with a lower-case letter, while predicates start with an upper-case letter. Axioms \eqref{axiom:dec_holdsAt} and \eqref{axiom:dec_holdsAt_inertia} express when a fluent holds, while axioms \eqref{axiom:dec_not_holdsAt} and \eqref{axiom:dec_not_holdsAt_inertia} denote the conditions in which a fluent does not hold. In CER, as we have formulated it here, the truth values of the composite events (CE)s of interest --- the `query atoms' --- are expressed by means of the $\folholdsAt$ predicate. The incoming 'simple, derived events' (SDE)s are represented by means of $\folhappensAt$, while any additional contextual information is represented by domain-dependent predicates. The SDEs and such contextual information constitute the `evidence atoms'. This way, CEs may be defined by means of $\folinitiatedAt$ and $\folterminatedAt$ predicates, stating the conditions in which a CE is initiated and terminated.

In order to learn Event Calculus theories, online structure learning methods may be employed in order to efficiently learn in the presence of data streams. \OSLa\ \citep{vagmcs2016osla} is an online structure learner, based on Markov Logic Networks (MLNs) \citep{domingos2006markov}, that can learn \MLNEC\ \citep{anskarlTOCL15} definitions --- a probabilistic variant of DEC --- by adapting the procedure of OSL \citep{huynh2011osl} and exploiting a given background knowledge. In particular, \OSLa\ exploits the \MLNEC\ axioms to constrain the space of possible structures during search. Each axiom contains $\folholdsAt$ predicates, that consist the supervision, and $\folinitiatedAt$, $\folterminatedAt$ predicates, that form the target CE definitions that we want to learn. \OSLa\ creates mappings from $\folholdsAt$ atoms to $\folinitiatedAt$, $\folterminatedAt$ atoms and searches only for explanations of the latter. Upon doing so, \OSLa\ only needs to find appropriate bodies over the current time-point to form clauses. Each incoming training example is used along the already learned clauses to predict the truth values of the $\folholdsAt$. Then, \OSLa\ constructs a hypergraph that represents the space of possible structures as graph paths. For all incorrectly predicted CEs the hypergraph is searched, using relational path-finding \citep{richards1992RP}, for clauses supporting the recognition of these incorrectly predicted CEs. The paths discovered during the search correspond to conjunctions of true ground evidence atoms (SDEs and contextual information) and are used to form clauses. The weights of the clauses that pass the evaluation stage are optimised using the AdaGrad online learner \citep{duchi2011AdaGrad}.

OLED \citep{KatzourisAP16} is based on Inductive Logic Programming, constructing CE definitions in the Event Calculus, in a single pass over the data stream. OLED constructs definitions by encoding each positive example, arriving in the input stream, into a so-called \emph{bottom rule}, i.e., a most specific rule of the form $\alpha \leftarrow \delta_1 \wedge \ldots \wedge \delta_n$, where $\alpha$ is an $\folinitiatedAt$ or $\folterminatedAt$ atom and $\delta_i$ are relational features (e.g., SDEs). A bottom clause is typically too restrictive to be useful, thus, OLED searches the space of all possible rules that $\theta$-subsume the bottom rule. To that end, OLED starts from the most-general rule and gradually specialises that rule, in a top-down fashion, by adding $\delta_i$'s to its body and using a rule evaluation function to assess the quality of each specialisation. OLED's single-pass strategy draws inspiration from the VFDT (Very Fast Decision Trees) algorithm \citep{DomingosH00} which is based on the Hoeffding bound, a statistical tool that allows to approximate the quality of a rule on the entire input using only a subset of the data. Thus, in order to decide between specialisations, OLED accumulates observations from the input stream until the difference between the best and the second-best specialisation satisfies the Hoeffding bound.

Both \OSLa\ and OLED have shortcomings. OLED is a crisp learner and therefore it cannot learn models that yield probabilistic inference capabilities. On the other hand, \OSLa\ is based on MLNs and thus inherits their probabilistic properties, but its structure learning component is sub-optimal, i.e., it tends to generate large sets of clauses, many of which have low heuristic value. An in-depth comparison of these systems can be found in \citep{KatzourisECML18}. More importantly, both \OSLa\ and OLED are supervised learners and in the presence of unlabelled training examples they impose closed-world assumption, that is, they assume everything not known is false, i.e., negative examples. This assumption can seriously compromise the learning task or even worse render it impossible if very little supervision is available, which is a common scenario in real-world applications.

\subsection{Harmonic Function Graph-Cut Minimisation} \label{sec:graph.cut}

Graph-based semi-supervised learning techniques \citep{ZhuSemiSupervised} construct a graph, whose vertices represent the labelled and unlabelled examples in the dataset and the edges reflect the similarity of these examples. Using such a graph, the learning task can be formulated as a graph-cut minimisation problem. The idea is to remove a minimal set of edges, so that the graph is cut into two disjoint sets of vertices; one holding positive examples and one holding negative ones.

Formally, let a training sequence consisting of $l$ labelled instances $\{(\mathbf{x}_i,\, y_i)\}_{i=1}^l$ and $u$ unlabelled ones $\{\mathbf{x}_j\}_{j=l+1}^{l+u}$. The labelled instances are pairs of a label $y_i$ and a $D$-dimensional numerical feature vector $\mathbf{x}_i = (x_1,\,\dots ,\, x_D) \in \mathbb{R}^D$ of input values, while the unlabelled ones are feature vectors with unknown label. Each of these instances represents either a labelled or an unlabelled vertex of the graph. These vertices are then connected by undirected weighted edges that encode their similarity according to a given distance function. Consequently, the labelled vertices can be used to determine the labels of the unlabelled ones. Once the graph is built, the task reduces into assigning $y$ values to the unlabelled vertices. Thus, the goal is to find a function $f(\mathbf{x}) \in \{ -1,\, 1\}$ over the vertices, where $-1$ is a negative label and $1$ a positive one, such that $f(\mathbf{x}_i)=y_i$ for labelled instances, and the cut size is minimised in order for the unlabelled ones to be assigned optimal values.

The minimum graph-cut can be represented as a regularised risk minimisation problem \citep{BlumMincut}, by using an appropriate loss function, forcing the labelled vertices to retain their values and a regularisation factor controlling the cut size. The cut size is the sum of the weights $w_{ij}$ corresponding to connected vertices $i$ and $j$ having different labels, and is computed as follows:

\begin{equation}\label{equation:cut.size}
	\sum_{i,j:\, f(\mathbf{x}_i) \neq f(\mathbf{x}_j)} w_{ij} = \sum_{i,j=1}^{l+u} w_{ij}\big(f(\mathbf{x}_i) - f(\mathbf{x}_j)\big)^2
\end{equation}

\noindent Equation \eqref{equation:cut.size} is an appropriate measure of the cut size, since it is affected only by edges for which $f(\mathbf{x}_i)\neq f(\mathbf{x}_j)$. Note that if $\mathbf{x}_i$ and $\mathbf{x}_j$ are not connected, then $w_{ij}=0$ by definition, while if the edge exists and is not cut, then $f(\mathbf{x}_i) - f(\mathbf{x}_j)=0$. Thus, the cut size is well-defined even when summing over all vertex pairs. Assuming that the maximal loss per edge is $R$, the loss for labelled instances should be zero if $f(\mathbf{x}_i)=y_i$ and $R$ otherwise. Thus, the loss function is defined as follows:

\begin{equation}\label{equation:loss}
	\ell\big(\mathbf{x}_i,\, y_i,\, f(\mathbf{x}_i)\big) = R\, \big(y_i - f(\mathbf{x}_i)\big)^2
\end{equation}

Consequently, by combining the loss function, as expressed by eq. \eqref{equation:loss} and the cut size, as expressed by eq. \eqref{equation:cut.size}, as a regularisation factor, the minimum graph-cut regularised risk problem is formulated as follows:

\begin{equation}\label{equation:minimum.cut}
\min_{ f:f(\mathbf{x}) \in \{ -1,\, 1 \} } 
R \sum_{i=1}^l \big(y_i - f(\mathbf{x}_i)\big)^2 +
\sum_{i,j=1}^{l+u} w_{ij} \big(f(\mathbf{x}_i) - f(\mathbf{x}_j)\big)^2
\end{equation}

Note that eq. \eqref{equation:minimum.cut} is an integer programming problem because $f$ is constrained to produce discrete values. Although efficient polynomial--time algorithms exist to solve the minimum graph-cut problem, still the formulation has a particular defect. There could be multiple equally good solutions; a label may be positive in one of the solutions, and negative in the rest. An alternative formulation proposed by \citep{ZhuGL03} for the graph-cut minimisation problem, that overcomes these issues, is based on the harmonic function. The proposed approach is based on harmonic energy minimisation of a Gaussian field and it has been shown to respect the harmonic property, i.e., the value of $f$ at each unlabelled vertex is the average of $f$ of the neighbouring vertices. In the context of semi-supervised learning, a harmonic function is a function that retains the values of the labelled data and satisfies the weighted average property on the unlabelled data:

\begin{equation} \label{eq:harmonic_function}
  \begin{aligned}
    f(\mathbf{x}_i) &= y_i,\, i=1,\dots, l\\
    f(\mathbf{x}_j) &= \frac{\sum_{k=1}^{l+u}w_{jk}f(\mathbf{x}_k)}{\sum_{k=1}^{l+u}w_{jk}},\, j=l{+}1,\dots, l{+}u
  \end{aligned}
\end{equation}

\noindent The former formula enforces that the labelled vertices retain their values, while the latter averages the labels of all neighbouring vertices of a given vertex, according to the weights of their edges. Therefore, the value assigned to each unlabelled vertex is the weighted average of its neighbours. The harmonic function leads to the same solution of the problem as defined in eq. \eqref{equation:minimum.cut}, except that $f$ is relaxed to produce real values. The main benefit of the continuous relaxation is that a unique optimal closed--form solution exists for $f$ that can be computed using matrix techniques. The drawback of the relaxation is that the solution is a real value in $[-1,\, 1]$ and does not directly correspond to a label. This issue can be addressed by thresholding $f$ at zero (\textit{harmonic threshold}) to produce discrete labels.

\subsection{Distance over Herbrand Interpretations} \label{sec:metric}

Distance functions constitute essential components of graph-based methods to semi-supervised learning and control the quality of the solution. In the case of numerical data, the Euclidean distance, the Gaussian kernel or Radial Basis Functions are common choices, as are matching distances for categorical data. However, in the presence of relational data there is a need for structure-based distances.

A technique proposed by \cite{NienhuysCheng} derives a distance for tree structure formalisms and thus provides a generic and natural approach for syntactic comparison of ground logical atoms. The distance function is defined on a set of expressions (namely ground atoms and ground terms), motivated by the structure and complexity of the expression, as well as the symbols used therein. Let $\mathcal{E}$ be the set of all expressions in a first-order language and $\mathbb{R}$ the set of real numbers. The distance $d: \mathcal{E} \times \mathcal{E} \mapsto \mathbb{R}$ over expressions $\mathcal{E}$, bounded by $1$, is defined as follows:

\begin{equation}\label{equation:distance}
\begin{array}{l}
	d(e,\, e) = 0,\, \forall e \in \mathcal{E} \\
	d(p(s_1,\, \dots ,\, s_k),\, q(t_1,\, \dots ,\, t_r)) = 1,\; p \neq q \vee k \neq r \\
	d(p(s_1,\, \dots ,\, s_k),\, q(t_1,\, \dots ,\, t_k)) = \frac{1}{2k}\sum_{i=1}^k d(s_i,\, t_i),\; p = q
\end{array}
\end{equation}

\noindent The first formula states that the distance of an expression to itself is zero. The second one states that if predicates $p$ and $q$ are not identical, either in terms of symbol or arity, then their distance is one because they refer to different concepts. We assume that the negation of a predicate $p$ has always distance $1$ from $p$, and thus, it can be seen as a special case of the second formula, where $q=\neg p$. In case $p$ and $q$ are identical, then their distance is computed recursively by the distance of the terms therein. The distance $d$ is also used by \cite{NienhuysCheng} over subsets of $\mathcal{E}$, i.e., sets of ground atoms, by means of the Hausdorff metric \citep{hausdorff1962set}. Informally, the Hausdorff metric is the greatest distance you can be forced to travel from a given point in one of two sets to the closest point in the other set.

A drawback of the Hausdorff metric \citep{Raedt2008,RamonB98} is that it does not capture much information about the two sets as it is completely determined by the distance of the most distant elements of the sets to the nearest neighbour in the other set. Thus, it may not be representative of the dissimilarity of the two sets. Formally, given the sets $\mathcal{E}_1$ and $\mathcal{E}_2$, their Hausdorff distance is computed as follows: 

\begin{equation*}
 \max\{ \sup_{x \in \mathcal{E}_1}\inf_{y \in \mathcal{E}_2} d(x,y), \sup_{y \in \mathcal{E}_2}\inf_{x \in \mathcal{E}_1} d(x,y) \}
\end{equation*}

The overall distance for these sets would be represented by one of the pairwise distances, namely the maximum distance among the minimum ones. Moreover, this type of approach allows one element in one set to match with multiple elements in the other set, which is undesirable because some elements may have no match and thus may be ignored in the resulting distance value. As stated by \cite{Raedt2008}, these limitations  motivate the introduction of a different notion of matching between two sets, which associate one element in a set to at most one other element. To that end, we employ the Kuhn-Munkres algorithm (Hungarian method) \citep{Kuhn1955}, which computes the optimal one-to-one assignment given some cost function, in our case the structural distance expressed by eq. \eqref{equation:distance}. The goal is to find the assignment of ground atoms among the sets that minimises the total cost, i.e., the total structural distance.

\section{Semi-Supervised Online Structure Learning} \label{sec:ssl}

Our goal is to effectively apply online structure learning in the presence of incomplete supervision. To do so, we take advantage of the structural dependencies underlying a logic-based representation and exploit regularities in the relational data, in order to correlate given labelled instances to unlabelled ones and reason about the actual truth values of the latter. Structure learning methods attempt to discover multi-relational dependencies in the input data, by combining appropriate evidence predicates, that possibly explain the given supervision, that is, the labelled ground query atoms of interest. The underlying assumption is that sets of ground evidence atoms that explain particular labelled query atoms are also contiguous to sets of ground evidence atoms that relate to unlabelled instances. One promising approach to model such similarities for partially supervised data is to use graph-based techniques. Graph-based methods attempt to formulate the task of semi-supervised learning as a graph-cut optimisation problem (see Section \ref{sec:graph.cut}) and then find the optimal assignment of values for the unlabelled instances given a similarity measure.

\begin{figure}[h]
\centering
\includegraphics[width=0.65\textwidth]{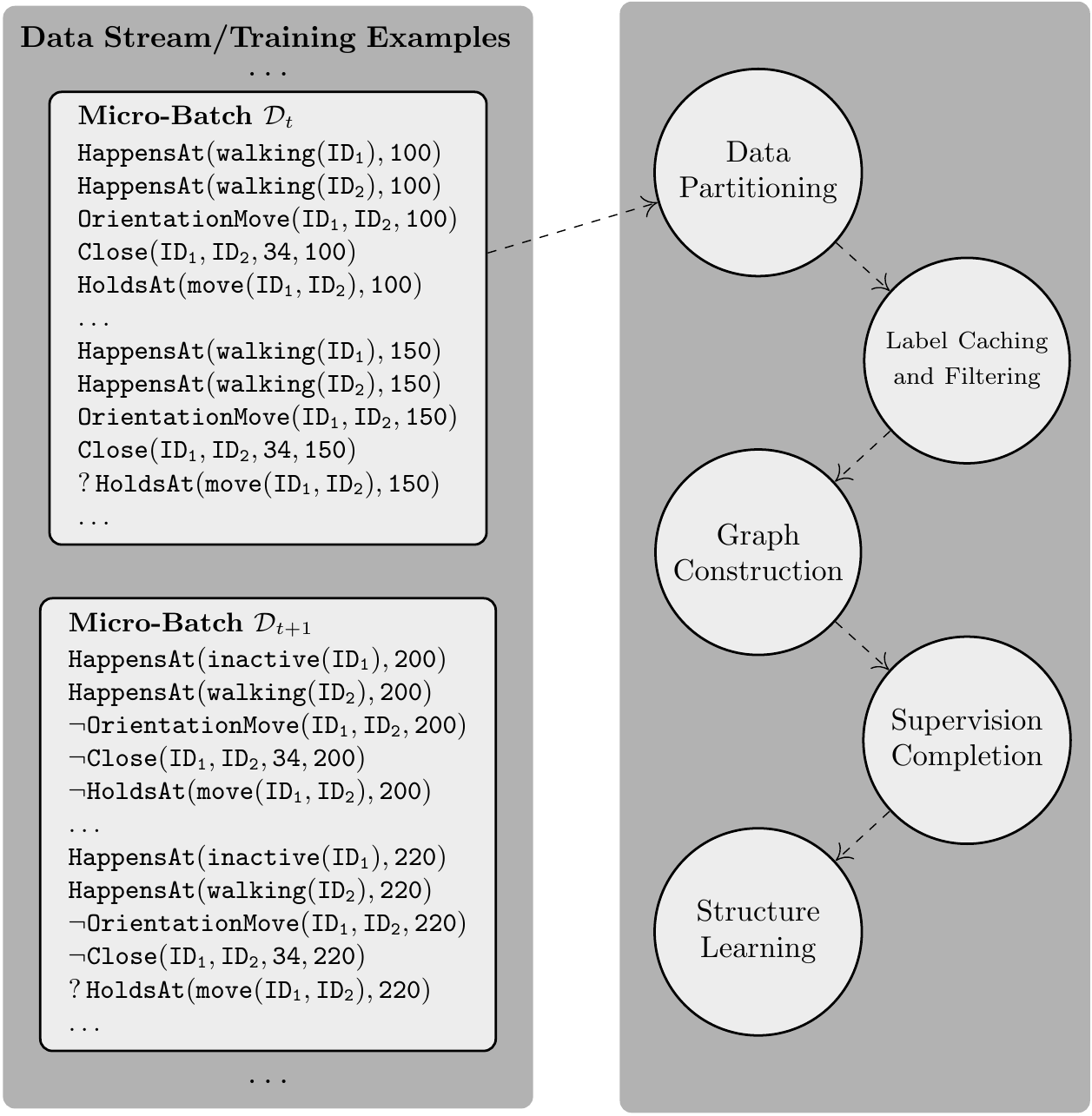}
\caption{The Semi-Supervised Online Structure Learning (SPLICE) procedure.}
\label{fig:splice}
\end{figure}

Figure \ref{fig:splice} presents the components and procedure of our proposed graph-based approach, using, for illustration purposes, the activity recognition domain as formalised in the Event Calculus. In order to address the online processing requirement, we assume that the training sequence arrives in micro-batches. At each step $t$ of the online procedure, a training example (micro-batch) $\mathcal{D}_t$ arrives containing a sequence of ground evidence atoms, e.g. two persons walking individually, their distance being less than $34$ pixel positions and having the same orientation. Each micro-batch may be fully labelled, partially labelled, or contain no labels at all. Labelling is given in terms of the Event Calculus $\folholdsAt$ query atoms. Unlabelled query atoms are prefixed by `?'. For instance, in micro-batch $\mathcal{D}_t$ there is no labelling for time-point $150$, while time-point $100$ expresses a positive label for the \move\ activity. Micro-batch $\mathcal{D}_t$ is passed onto the data partitioning component that groups the training sequence into examples. Each unique labelled example present in the micro-batch is stored in a cache, in order to be reused in subsequent micro-batches that may have missing labels. 

Labelled and unlabelled examples are converted into graph vertices, linked by edges that represent their structural similarity. The resulting graph is then used to label all unlabelled ground query atoms. Given the fully labelled training sequence, an online structure learning step refines or enhances the current hypothesis --- and the whole procedure is repeated for the next training micro-batch $\mathcal{D}_{t+1}$. For the online structure learning component we may use \OSLa\ or OLED (see Section \ref{sec:ec_learning}). 

Henceforth, we refer to our proposed approach as SPLICE (\textbf{s}emi-su\textbf{p}ervised on\textbf{li}ne stru\textbf{c}ture l\textbf{e}arning). The components of our method are detailed in the following subsections. To aid the presentation, we use examples from human activity recognition.

\begin{figure}[t]
\begin{center}
\includegraphics[width=0.8\textwidth]{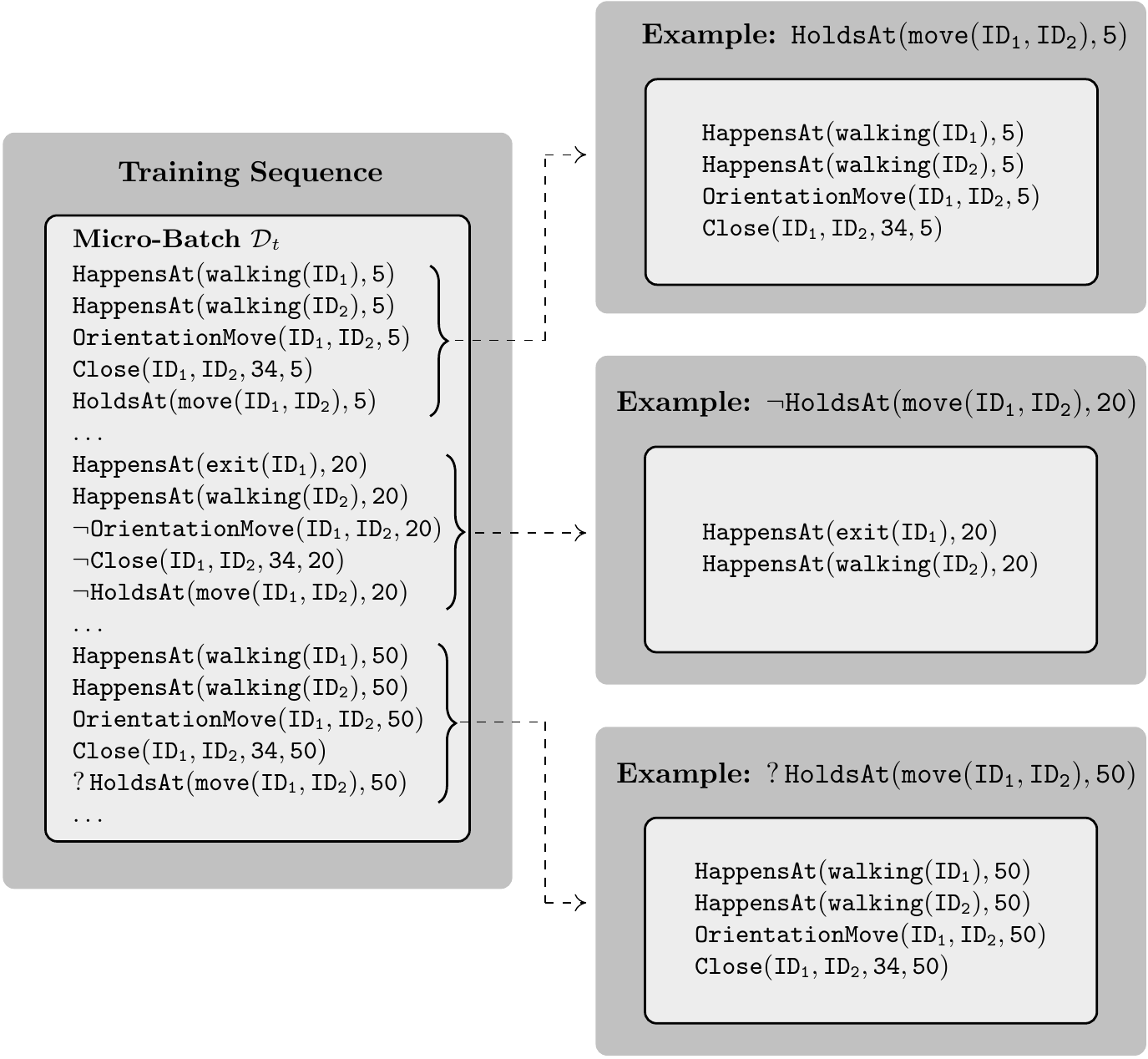}
\end{center}
\caption{Data partitioning into examples. Each example contains a ground query atom, either labelled or unlabelled, as well as a set of true ground evidence atoms that relate to the query atom through their constants.}
\label{fig:partition}
\end{figure}

\subsection{Data Partitioning} \label{sec:partition}

In a typical semi-supervised learning setting, the training sequence consists of both labelled instances $\{\mathbf{x}_i,\, y_i\}_{i=1}^l$ and unlabelled ones $\{\mathbf{x}_j\}_{j=l+1}^u$ where each label $y_i$ corresponds to a $D$-dimensional feature vector $\mathbf{x}_i = (x_1,\,\dots ,\, x_D) \in \mathbb{R}_D$ of input values. Given a logic-based representation of instances, our approach begins by partitioning the given input data (micro-batch $\mathcal{D}$) into sets of ground evidence atoms, each one connected to a supervision (query) ground atom. The resulting sets are treated as training examples. Let $\mathcal{E} = \{e_1,\,\dots,\, e_M\}$ be the set of all true evidence ground atoms and $\mathcal{Q} = \{q_1,\,\dots,\, q_N\}$ the set of all ground query atoms of interest in micro-batch $\mathcal{D}$. Each example should contain exactly one ground query atom $q_i$ and a proper subset $\mathcal{E}_i \subset \mathcal{E}:\, i=\{1,\,\dots ,\, N\}$ of evidence atoms corresponding to $q_i$. Given the sets $\mathcal{E}$ and $\mathcal{Q}$, we construct an example for each ground query atom in $\mathcal{Q}$, regardless whether it is labelled or not. To do so, we partition the evidence atoms in $\mathcal{E}$ into non-disjoint subsets, by grouping them over the constants they share directly to the ground query atom $q_i$ of each example. A constant is shared if and only if it appears in both atoms and its position on both atoms has the same type. Note that the position of a constant in some evidence atom $e$ may differ from that in $q_i$. We refrained from including longer range dependencies, such as considering evidence atoms that can be reached through several shared constants, to favour run-time performance.

Figure \ref{fig:partition} illustrates the presented procedure. As usual, $\folholdsAt$ express query atoms, while all other predicates express evidence atoms. Unlabelled query atoms are denoted by the prefix `?'. Data partitioning takes into account only true evidence atoms and concerns only a specific query predicate. Note that each resulting example has a set $\mathcal{E}_i \subset \mathcal{E}$ of evidence atoms that comprise only constants relevant to the query atom. For instance, the ground evidence atom $\mathtt{Close(ID_1,ID_2,34,5)}$ appearing only in the top example, shares constants $\mathtt{ID_1,ID_2}$ with query atoms of other examples too, but constant $\mathtt{5}$ is only relevant to the top example. Constant $\mathtt{34}$ does not appear in any query atom and thus can be ignored. Similarly, ground evidence atoms having constants that appear in many query atoms will appear in all corresponding examples. This is an expected and desirable behaviour, because such predicates indeed capture knowledge that may be important to many query atoms. For instance, consider a ground predicate $\mathtt{Person(ID_1)}$ stating that $\mathtt{ID_1}$ is a person. If such a predicate was included in the evidence of Figure \ref{fig:partition}, it would appear in every example. Moreover, during data partitioning, SPLICE can ignore specific predicates according to a set of given mode declarations \citep{muggleton95}, using the recall number. If the recall number is zero the predicate is ignored.

\begin{algorithm}[h]
\caption{Partition($\mathcal{D}$, md)}
\label{alg:data_partitioning}

\begin{algorithmic}[1]
\Require{$\mathcal{D}$: a training micro-batch, md: a set of mode declarations}
\Ensure{$\mathcal{V}$: a set of vertices}
\State{Partition $\mathcal{D}$ into $\mathcal{Q}$ and $\mathcal{E}$.}
\State{$\mathcal{V}=\emptyset$}
\Statex{\# $c_{q1},\dots, c_{qn}$ are constants}
\ForAll{ground query atoms $q(c_{q1},\dots,c_{qn}) \in \mathcal{Q}$}
    \State{$\mathcal{E}_q=\emptyset$}
    \ForAll{true ground evidence atoms $e(c_{e1},\dots,c_{em})\in \mathcal{E} : \mathrm{recall} > 0$}
      \State{$C_{e,q} = \{c \, | \, c \in e \land \mathrm{type}(c,e) \in \mathrm{Types}(e) \cap  \mathrm{Types}(q)\}$}
      \If{$\forall\; c_i \in C_{e,q} \; \exists\; c_j \in q : c_i = c_j \land \mathrm{type}(c_i,e) = \mathrm{type}(c_j,q)$}
          \State{$\mathcal{E}_q=\mathcal{E}_q \cup e(c_{e1},\dots,c_{em})$}
      \EndIf
    \EndFor
    \State{$\mathcal{V}=\mathcal{V} \cup \{(q(c_{q1},\dots,c_{qn}),\, \mathcal{E}_q)\}$}
\EndFor\\

\Return{$\mathcal{V}$}
\end{algorithmic}
\end{algorithm}

We henceforth refer to examples as vertices, since each example is represented by a vertex in the graph which is subsequently used by the graph-cut minimisation process. Algorithm \ref{alg:data_partitioning} presents the pseudo-code for partitioning input data into examples representing the graph vertices. The algorithm requires as an input a training micro-batch $\mathcal{D}$ and a set of mode declarations, and produces a set of vertices. At line $1$ the micro-batch is partitioned into a set of ground query atoms $\mathcal{Q}$ and a set of ground evidence atoms $\mathcal{E}$. Then at line $3$ the algorithm iterates over all ground query atoms and for each one it finds all true ground evidence atoms sharing constants of the same type. The set $C_{e,q}$ includes a constant $c$ of an evidence atom $e$ if and only if the position of $c$ in $e$ has type $t$, and $t$ is present in the query atom $q$. Then, $e$ is added to the vertex of $q$ if all constants of $C_{e,q}$ appear in $q$, and their positions on both $e$ and $q$ have the same type. Function $\mathrm{type}(c,a)$, appearing in line $6$, gives the type of the position of constant $c$ in atom $a$, while $\mathrm{Types}(a)$ gives all the types of $a$. Finally, for each pair of ground query atom and its corresponding set of relevant ground evidence atoms, the algorithm creates a vertex and appends it to the vertex set.

\subsection{Graph Construction} \label{sec:graph_construction}

Once the example vertices have been constructed, supervision completion assigns truth values to the unlabelled vertices, by exploiting information present in the labelled ones. A weighted edge between a particular pair of vertices $v_i,\, v_j: i, j \in \{1,\dots,N\}$ represents the structural similarity of the underlying ground evidence atom sets in the two vertices. Note that the number of vertices is equal to the number of ground query atoms in $\mathcal{Q}$, that is $N$. Let $w_{i j}$ be the edge weight, i.e., the structural similarity of $v_i$ and $v_j$. If $w_{i j}$ is large enough, then the truth values of the ground query atoms $q_i,\, q_j$ are expected to be identical. Therefore, the similarity measure essentially controls the quality of the supervision completion solution.

Our approach regarding the computation of the evidence atom similarities is based on a measure of structural dissimilarity $d: \mathcal{E} \times \mathcal{E} \mapsto \mathbb{R}$ over a set of first-order expressions $\mathcal{E}$. The distance $d$ does not make any syntactical assumptions about the expressions, such as function-free predicates, and thus is applicable to any domain of interest. As described in Section \ref{sec:metric}, we define a measure over sets of ground atoms using the Kuhn-Munkres algorithm, which provides an optimal one-to-one mapping given a cost matrix. In our case the cost matrix essentially holds the distances between each pair of ground atoms, computed by eq. \eqref{equation:distance}, present in the sets being compared. In particular, for each pair of vertices $v_i = (\mathcal{E}_i,\, q_i),\, v_j = (\mathcal{E}_j,\, q_j)$ our approach begins by computing the distance between each pair of expressions $d(e_{i,k},\, e_{j,l}): e_{i,k} \in \mathcal{E}_i,\, e_{j,l} \in \mathcal{E}_j$ resulting in a matrix $\mathbf{C}$ that represents the costs of the assignment problem:

\begin{equation*}
\mathbf{C} = 
 \begin{pmatrix}
  d(e_{i,1},\, e_{j,1}) & d(e_{i,1},\, e_{j,2}) & \cdots & d(e_{i,1},\, e_{j,M}) \\
  d(e_{i,2},\, e_{j,1}) & d(e_{i,2},\, e_{j,2}) & \cdots & d(e_{i,2},\, e_{j,M}) \\
  \vdots  & \vdots  & \ddots & \vdots  \\
  d(e_{i,M},\, e_{j,1}) & d(e_{i,M},\, e_{j,2}) & \cdots & d(e_{i,M},\, e_{j,M})
 \end{pmatrix}
\end{equation*}

\noindent This matrix is square $M \times M$, assuming that the sets $\mathcal{E}_i$ and $\mathcal{E}_j$ are of equal size. In the general case, of a $M \times K$ matrix, where $M > K$, $\mathbf{C}$ is padded using zero values to complete the smaller dimension in order to be made square. Intuitively, the zero values in the smaller set capture the notion of unmatched atoms. The matrix $\mathbf{C}$ can then be used as the input cost matrix for the Kuhn-Munkres algorithm, in order to find the optimal mapping of evidence atoms. 

The optimal mapping is denoted here by the function $g: \mathbb{R}^{M \times M} \mapsto \{(i,\, j):\, i,\,j \in \{1,\, \dots ,\, M\} \}$ and is the one that minimises the total cost, i.e., the sum of the distances of the mappings, normalised by the dimension $M$ of the matrix:

\begin{equation}\label{eq:cost}
	cost = \min_{g} \frac{1}{M} \Big( (M - K) + \sum_{g(\mathbf{C)} \mapsto (i,j)} \mathbf{C}_{i,j} \Big)
\end{equation}

The unmatched evidence atoms constitute an important component of the total cost, due to the term $M-K$, which penalises every unmatched ground atom by the greatest possible distance, that is $1$. Thus, $M-K$ can be seen as a regularisation term. The need to penalise unmatched atoms stem from the fact that they may represent important features that discriminate a positive from a negative example. The normalised total cost is translated into a similarity $z(v_i, v_j) = 1 - cost$ and assigned as the weight $w_{i j}$ of the edge connecting the vertices $v_i,\, v_j$. The measure denoted by the function $z$ is symmetric and is used to calculate the similarity of all distinct vertex pairs. The process generates a $N \times N$ symmetrical adjacency matrix $\mathbf{W}$ comprising the weights of all graph edges. Hence, matrix $\mathbf{W}$ is computed using eq. \eqref{eq:cost} through function $z$. To avoid self-loops, i.e., edges that connect a vertex to itself, we set the diagonal of the $\mathbf{W}$ matrix to zero:

\begin{equation*}
\mathbf{W} =
 \begin{pmatrix}
  0 & z(v_1,\, v_2) & \cdots & z(v_1,\, v_N) \\
  z(v_2,\, v_1) & 0 & \cdots & z(v_2,\, v_N) \\
  \vdots  & \vdots  & \ddots & \vdots  \\
  z(v_N,\, v_1) & z(v_N,\, v_2) & \cdots & 0 \\
 \end{pmatrix}
\end{equation*}

In order to turn the similarity matrix $\mathbf{W}$ into a graph, we use a connection heuristic, which introduces edges only between vertices that are very similar, i.e., they have a high weight. In the simplest case, we connect the vertices $v_i,\, v_j$ if $z(v_i,\, v_j) \geq \epsilon$, given some threshold value $\epsilon$ ($\epsilon$NN). Another alternative is to use $k$ nearest neighbour ($k$NN) to choose the edges that will be kept. According to this approach, for each vertex $v_i$ we identify the closest (most similar) $k$ vertices. Note that if $v_i$ is among $v_j$'s $k$ nearest neighbours, the reverse is not necessarily true. In order to avoid tie-breaking, we modified $k$NN to select the top $k$ distinct weights in a vertex neighbourhood, and then connect all neighbours having such a weight.

\subsection{Supervision Completion} \label{sec:completion}

Given the weight matrix $\mathbf{W}$, we apply one of the two connection heuristics mentioned above to obtain a sparse matrix $\mathbf{W}'$, having zeros for unconnected vertices and a similarity value $w$ for the rest. Matrix $\mathbf{W}'$ is used to perform graph-cut minimisation to assign truth values to the unlabelled ground query atoms. 

Let $l+u = N$ be the number of labelled and unlabelled vertices. The closed-form solution of the optimisation problem for the harmonic function (see Section \ref{sec:graph.cut}) in matrix notation is as follows. Let $D_{ii}$ be the weighted degree of vertex $i$, i.e., the sum of the edge weights connected to $i$. Let $\mathbf{D}$ be a $N \times N$ diagonal matrix, containing $D_{ii}$ on the diagonal, computed over the matrix $\mathbf{W}'$. Then the unnormalised graph Laplacian matrix $\mathbf{L}$ is defined as follows:

\begin{equation*}
	\mathbf{L} = \mathbf{D} - \mathbf{W}'
\end{equation*}

In this case, the Laplacian matrix essentially encodes the extent to which the harmonic function $f$ (see eq. \eqref{eq:harmonic_function}) differs at a vertex from the values of nearby vertices. Assuming that vertices are ordered so that the labelled ones are listed first, the Laplacian matrix can be partitioned into four sub-matrices as follows:

\begin{equation*}
\mathbf{L} = 
\begin{bmatrix}
  \mathbf{L}_{ll} & \mathbf{L}_{lu} \\
  \mathbf{L}_{ul} & \mathbf{L}_{uu}
\end{bmatrix}
\end{equation*}

\noindent The partitioning is useful in order to visualise the parts of $\mathbf{L}$. Sub-matrices $\mathbf{L}_{ll},\mathbf{L}_{lu}, \mathbf{L}_{ul}$ and $\mathbf{L}_{uu}$ comprise, respectively, the harmonic function differences between labelled vertices, labelled to unlabelled, unlabelled to labelled and unlabelled to unlabelled. Note that $\mathbf{L}_{lu}$ and $\mathbf{L}_{ul}$ are symmetric.

Let $\mathbf{f} = (f(\mathbf{x}_1),\, \dots ,\, f(\mathbf{x}_{l+u}))^{\top}$ be the vector of $f$ values of all vertices and the partitioning of $\mathbf{f}$ into $(\mathbf{f}_l,\, \mathbf{f}_u)$ hold the values of the labelled and unlabelled vertices respectively. Then by solving the constrained optimisation problem, expressed in eq. \eqref{equation:minimum.cut}, using the Lagrange multipliers and matrix algebra, one can formulate the harmonic solution as follows:

\begin{equation} \label{equation:minimum.cut.closed}
  \begin{aligned}
    \mathbf{f}_l &= \mathbf{y}_l\\
    \mathbf{f}_u &= -\mathbf{L}_{uu}^{-1}\mathbf{L}_{ul}\mathbf{y}_l
  \end{aligned}
\end{equation}

\noindent Since $\mathbf{L}_{lu}$ and $\mathbf{L}_{ul}$ are symmetric, any of the two can be used to solve the optimisation defined but eq. \eqref{equation:minimum.cut.closed}. However, if we use $\mathbf{L}_{lu}$ instead of $\mathbf{L}_{ul}$, then its transpose should be used in order for the matrix dimensions to agree during the multiplications. Equation \eqref{equation:minimum.cut.closed} requires the computation of the inverse of matrix $\mathbf{L}_{uu}$ that may be singular, due to many zero values (sparsity). In order to avoid this situation, we replace zeros by a very small number. A different solution would be to use the pseudo-inverse, but the computation proved significantly slower in our datasets, without significant differences in accuracy. Since the optimal solution is required to comprise the labels assigned to unlabelled vertices in $[-1,\, 1]$, the resulting solution $\mathbf{f}_u$ is thresholded at zero to produce binary labels\footnote{We also experimented with adaptive threshold approaches designed to handle the possible class imbalance by exploiting the class prior probabilities. We tried a threshold based on log-odds and an approach proposed by \cite{ZhuGL03}, called class mass normalisation. Both of them yielded much worse results than the harmonic threshold in our experiments.}.

\begin{algorithm}
\caption{SupervisionCompletion($\mathcal{V}$, $h$, $z$)}
\label{alg:supervision_completion}

\textbf{Input:} $h$: a connection heuristic, $z$: structural similarity,\\
$\mathcal{V}$: a set of labelled and unlabelled vertices\\
\textbf{Output:} $\mathbf{f}_u$: labels for the unlabelled query atoms

\begin{algorithmic}[1]
\State{Initialise matrix $\mathbf{W}$ to be the zero matrix $\mathbf{0}$}
\For{$v_i \in \mathcal{V}$}
    \For{$v_j \in \mathcal{V}$}
        \State{$w_{i,j} = z(v_i,v_j)$}
    \EndFor
\EndFor

\State{Apply the connection heuristic: $\mathbf{W}' = h(\mathbf{W})$}
\State{Compute Laplacian matrix: $\mathbf{L} = \mathbf{D} - \mathbf{W}'$}
\State{Compute the graph-cut minimisation: $\mathbf{f}_u = -\mathbf{L}_{uu}^{-1}\mathbf{L}_{ul}\mathbf{y}_l$}
\Statex{\# Perform thresholding to acquire binary labels}
\For{$f_i \in \mathbf{f}_u$}
    \IfSingle{$f_i <$ small value} $f_i = -1$ which represents false
    \ElseSingle $f_i = 1$ which represents true
\EndFor\\

\Return{$\mathbf{f}_u$}
\end{algorithmic}
\end{algorithm}

Algorithm \ref{alg:supervision_completion} presents the pseudo-code for constructing the graph and performing supervision completion. The algorithm requires as input a connection heuristic, a structural similarity and a set of vertices, and produces as output a set of labels for the unlabelled vertices. First, we compute the similarity between all pairs of vertices (see lines $1$--$4$). Then we apply the connection heuristic to the matrix $\mathbf{W}$ holding the similarity values, compute the Laplacian matrix and solve the optimisation problem (see lines $5$--$7$). Finally, for the resulting vector $\mathbf{f}_u$ holding the values of the unlabelled vertices, we perform thresholding on each value yielding binary labels (see lines $8$-$10$). Since unlabelled examples are typically much more than the labelled ones (in a micro-batch), the inversion of the Laplacian matrix, yielding time $|\mathcal{Q}_U|^3$, is the main overhead of the algorithm, where $|\mathcal{Q}_U|$ denotes the number of unlabelled ground query atoms in a micro-batch\footnote{The complexity analysis of all steps of SPLICE may be found at:\\\url{https://iit.demokritos.gr/~vagmcs/pub/splice/appendix.pdf}}.

\subsection{Label Caching and Filtering} \label{sec:cache_filter}

In order to handle real-world applications where labelled examples are infrequent, our method --- SPLICE --- uses a caching mechanism, storing previously seen labelled examples for future usage. At each step of the online supervision completion procedure, SPLICE stores all unique labelled examples that are not present in the cache and then uses the cached examples to complete the missing labels. For each labelled vertex it creates a clause, using the label as the head, the true evidence atoms as the body, and replacing all constants with variables according to a set of given mode declarations \citep{muggleton95}. For instance, the second vertex of Figure \ref{fig:partition} can be converted to the following clause:

\begin{equation} \label{logic:clause}
  \begin{aligned}
    & \neg \mathtt{HoldsAt(move(id_1,id_2), t)}\, \Leftarrow \\
    & \mathtt{HappensAt(exit(id_1), t)}\, \wedge\, \mathtt{HappensAt(walking(id_2), t)}
  \end{aligned}
\end{equation}

\noindent For each such clause, SPLICE checks the cache for stored vertices that represent identical clauses and selects only the unique ones. The unique cached vertices are then used as labelled examples in the graph construction process of supervision completion in the current and subsequent micro-batches.

In any (online) learning task, noise, such as contradicting examples, is a potential risk that may compromise the accuracy of the learning procedure. In order to make SPLICE tolerant to noise, we use the Hoeffding bound \citep{hoeffding1963}, a probabilistic estimator of the error of a model (true expected error), given its empirical error (observed error on a training subset) \citep{DhurandharD12}. Given a random variable $X$ with a value range in $[0,1]$ and an observed mean $\bar{X}$ of its values after $N$ independent observations, the Hoeffding bound states that with probability $1-\delta$ the true mean $\mathit{\mu_X}$ of the variable lies in an interval $(\bar{X}-\varepsilon, \bar{X}+\varepsilon)$, where $\varepsilon = \sqrt{\ln(2/\delta)/2N}$. In other words, the true average can be approximated by the observed one with probability $1-\delta$ given an error margin $\varepsilon$.

In order to remove noisy examples, we detect contradictions in the cached labelled vertices, using the subset of training data that has been observed so far in the online process. To do so, we use an idea proposed by \cite{DomingosH00}. Let $c$ be the clause of a cached vertex $v$ and $n_c$ the number of times the clause has appeared in the data so far. Recall that the clause of a cached vertex is lifted, i.e. all constants are replaced by variables. Thus lifted clauses may appear many times in the data. Similarly, let $c'$ be the opposite clause of $c$, that is, a clause having exactly the same body but a negated head, and $n_{c'}$ its counts. For instance the opposite clause of \eqref{logic:clause} is:

\begin{equation*}
  \begin{aligned}
    & \mathtt{HoldsAt(move(id_1,id_2), t)}\, \Leftarrow \\
    & \mathtt{HappensAt(exit(id_1), t)}\, \wedge\, \mathtt{HappensAt(walking(id_2), t)}
  \end{aligned}
\end{equation*}

\noindent The goal is to eventually select only one of the two contradicting clauses. We define a function $p(c)=\frac{n_c}{n_c+n_{c'}}$ with range in $[0,1]$ that represents the probability of clause $c$ to appear in the data instead of its opposite clause $c'$. Then according to the Hoeffding bound, for the true mean of the probability difference $\Delta p = p(c)-p(c')$ it holds that $\Delta \bar{p} - \varepsilon < \Delta p$, with probability $1-\delta$. Hence, if $\Delta \bar{p} > \varepsilon$, we accept the hypothesis that $c$ is indeed the best clause with probability $1-\delta$ and thus $v$ is kept at this point. Similarly, $c'$ is the best one if $-\Delta \bar{p} > \varepsilon$. Therefore, in order to select between contradicting labelled examples, it suffices to accumulate observations until their probability difference exceeds $\varepsilon$. Until that point both example vertices are used in the optimisation.

\begin{algorithm}
\caption{CacheUpdateAndFilter($\mathcal{V}_L$, $\mathcal{C}$)}
\label{alg:cache_update}

\textbf{Input:} $\mathcal{V}_L$: a set of labelled vertices,\\
$\mathcal{C}$: cache containing a list of vertices and their counts\\
\textbf{Output:} $\mathcal{V}_L'$: a set of filtered labelled vertices, $\mathcal{C}$: the updated cache

\begin{algorithmic}[1]

    \For{$v_i \in \mathcal{V}_L$}
        \If{$\exists\; v_j \in \mathcal{C} : \mathrm{canUnify}(\mathrm{clause}(v_i), \mathrm{clause}(v_j))$}
            \State{$\mathcal{C}[v_j] = \mathcal{C}[v_j] + 1$}
        \Else
            \State{$\mathcal{C}[v_i] = 1$}
        \EndIf
    \EndFor
    
    \State{Initialise accumulated unique filtered labelled nodes $\mathcal{V}_L' = \emptyset$}
    \For{$(v_i,\, n) \in \mathcal{C}$}
        \State{Generate clause $c = \mathrm{clause}(v_i)$ and its opposite $c'$}
        \If{$\exists\; v_j \in \mathcal{C}: \mathrm{clause}(v_j) = c'$}
            \State{Compute total number of groundings $N = \mathcal{C}[v_i] + \mathcal{C}[v_j]$} 
            \State{Compute frequencies $p_c = \frac{\mathcal{C}[v_i]}{N}, p_{c'} = \frac{\mathcal{C}[v_j]}{N}$}
            \State{Compute $\varepsilon=\sqrt{\frac{\ln(2/\delta)}{2N}}$}
            \If{$p_c - p_{c'} > \varepsilon$} 
                \State{$\mathcal{V}_L' = \mathcal{V}_L' \cup v_i$}
            \EndIf
        \Else
            \State{$\mathcal{V}_L' = \mathcal{V}_L' \cup v_i$}
        \EndIf
    \EndFor\\

    \Return{$\mathcal{V}_L',\, \mathcal{C}$}
\end{algorithmic}
\end{algorithm}

Although we use the Hoeffding inequality to make the best filtering decision for contradicting examples, given the data that we have seen so far, the examples are not independent as the Hoeffding bound requires. Consequently, we allow this filtering decision to change in the future, given the new examples that stream-in, by keeping frequency counts of the lifted examples\footnote{We assume that the examples stem from a stationary stochastic process and thus the difference between contradicting example frequencies eventually converges when a sufficient amount of observations is accumulated.}. This is not the case in other applications \citep{DomingosH00,AbdulsalamSM11} in which the decision is permanent.

Algorithm \ref{alg:cache_update} presents the pseudo-code for cache update and filtering. The algorithm requires as input the labelled vertices of the current micro-batch and the cached vertices along with their counts, and produces as output the set of filtered labelled vertices and the updated cache. If the clause view of a vertex exists in the cache then the counter of that vertex is incremented, otherwise the vertex is appended in the cache and its counter is set to $1$ (see lines $1$--$5$). For each vertex in the cache we produce its clause and check if the cache contains a vertex representing the opposite clause. In case the opposite clause exists, the Hoeffding bound is calculated in order to check if one of them can be filtered out (see lines $6$--$16$). In the case that many labelled examples have been accumulated in the cache, update and filtering can have an impact on performance, yielding a total time of $t^2|\mathcal{Q}_U|^2 $, where $t$ is the number of micro-batches seen so far, and $|\mathcal{Q}_U|$ is the number of unlabelled query atoms in a micro-batch. Algorithm \ref{alg:splice} presents the complete SPLICE procedure.

\begin{algorithm}
\caption{SPLICE($h$, $z$, $\delta$, md)}
\label{alg:splice}

\textbf{Input:} $h$: connection heuristic, $z$: structural similarity,\\[0.1cm]
$\delta$: Hoeffding bound confidence, md: Mode declarations

\begin{algorithmic}[1]
\State{Initialise cache containing list of vertices and their counts $\mathcal{C} = \emptyset$}
\For{$t=1$ to $I$ micro-batches}
    \State{Receive a micro-batch $\mathcal{D}_t = (\mathcal{Q}_t,\mathcal{E}_t)$}
    \Statex{\# $\mathcal{Q}_t$ is a set of ground query atoms and $\mathcal{E}_t$ a set of ground evidence atoms.}
    \State{Partition data into vertices $\mathcal{V}=\mathbf{Partition}(\mathcal{D}_t, \mathrm{md})$}    
    \State{Partition $\mathcal{V}$ into labelled $\mathcal{V}_L$ and unlabelled $\mathcal{V}_U$ vertices}
    \State{$\mathcal{V}_L',\, \mathcal{C}=\mathbf{CacheUpdateAndFilter}(\mathcal{V}_L, \mathcal{C})$}

    \State{Union of the unique labelled nodes with unlabelled ones: $\mathcal{V}'{=}\mathcal{V}_L' \cup \mathcal{V}_U$}

    \State{$\mathbf{f}_u=\mathbf{SupervisionCompletion}(\mathcal{V}', h, z)$}
    \State{Perform a structure learning step using $(\mathbf{f}_l,\mathbf{f}_u)$}
\EndFor
\end{algorithmic}
\end{algorithm}


\section{Empirical Evaluation} \label{sec:evaluation}

We evaluate SPLICE on the task of composite event recognition (CER), using \OSLa\ and OLED as the underlying structure learners (see Section \ref{sec:ec_learning}). We use the publicly available benchmark video surveillance dataset of the CAVIAR project\footnote{\url{http://homepages.inf.ed.ac.uk/rbf/CAVIARDATA1}}, as well as a real maritime surveillance dataset provided by the French Naval Academy Research Institute (NARI), in the context of the datAcron project\footnote{\url{http://datacron-project.eu/}}.

\subsection{Experimental Setup} \label{sec:setup}

The video surveillance dataset comprises $28$ surveillance videos, where each video frame is annotated by human experts on two levels. The first level contains SDEs that concern activities of individual persons, such as when a person is walking or staying inactive. Additionally, the coordinates of tracked persons are also used to express qualitative spatial relations, e.g. two persons being relatively close to each other. The second level contains CEs, describing the activities between multiple persons and/or objects, i.e., people meeting and moving together, leaving an object and fighting. Similar to earlier work \citep{anskarlTOCL15,KatzourisAP16}, we focus on the \meet\ and \move\ CEs, and from the $28$ videos, we extract $19$ sequences that contain annotation for these CEs. The rest of the sequences in the dataset are ignored, as they do not contain positive examples of these two target CEs. Out of the $19$ sequences, $8$ are annotated with both \meet\ and \move\ activities, $9$ are annotated only with \move\, and $2$ only with \meet. The total length of the extracted sequences is $12{,}869$ video frames. Each frame is annotated with the (non-)occurrence of a CE and is considered an example instance. The whole dataset contains a total of $63{,}147$ SDEs and $25{,}738$ annotated CE instances. There are $6{,}272$ example instances in which \move\ occurs and $3{,}722$ in which \meet\ occurs. Consequently, for both CEs the number of negative examples is significantly larger than the number of positive ones.

The maritime dataset ($\approx 1.2$GiB) consists of position signals from $514$ vessels sailing in the Atlantic Ocean, around Brest, France. The SDEs express compressed trajectories in the form of `critical points', such as communication gap (a vessel stops transmitting position signals), vessel speed change, and turn. It has been shown that compressing vessel trajectories in this way allows for accurate trajectory reconstruction, while at the same time improving stream reasoning times significantly \citep{PatroumpasAAVPT17}. The dataset contains a total of $16{,}152{,}631$ SDEs. We focus on the \rendezvous\ CE, where two vessels are moving slowly in the open sea and are close to each other. Since the dataset is unlabelled, we produced synthetic annotation by performing CER using the RTEC engine \citep{ArtikisSP15} and a hand-crafted definition of \rendezvous. This way, \rendezvous\ occurs at $735{,}600$ out of the $6{,}832{,}124$ time-points of the dataset.

Throughout the experimental analysis, the accuracy results for both supervision completion and structure learning were obtained using the $F_1$-score. All reported statistics are micro-averaged over the instances of CEs. For the CAVIAR dataset, the reported statistics for structure learning were collected using $10$-fold cross validation over the $19$ video sequences, while complete videos were left out for testing. In the maritime dataset, the statistics were collected using $10$-fold cross validation over one month of data, while pairs of vessels were left out for testing. The experiments were performed in a computer with an Intel i7 4790@3.6GHz CPU ($4$ cores, $8$ threads) and $16$GiB of RAM. SPLICE and \OSLa\ are included in LoMRF\footnote{\url{https://github.com/anskarl/LoMRF}}, an open-source framework for MLNs, and OLED is available as an open-source ILP solution\footnote{\url{https://github.com/nkatzz/OLED}}. All presented experiments are reproducible\footnote{Instructions for reproducing all presented experiments can be found in:\\ \url{https://iit.demokritos.gr/~vagmcs/pub/splice}}.

\subsection{Hyperparameter Selection} \label{sec:selection}

We ran supervision completion on the CAVIAR dataset, for five values of $k$ and $\epsilon$, controlling the $k$NN and $\epsilon$NN connection heuristics (see Section \ref{sec:graph_construction}), in order to select the best configuration. Each micro-batch retained a percentage of the given labels, selected uniformly. We used $5,\, 10,\, 20,\, 40$ and $80\%$ supervision levels for the micro-batches, retaining the corresponding proportion of the labels. We repeated the uniform selection $20$ times, leading to $20$ datasets per supervision level, in order to obtain a good estimate of the performance of the method.

\begin{figure}[h]
\centering
\includegraphics[width=0.49\textwidth]{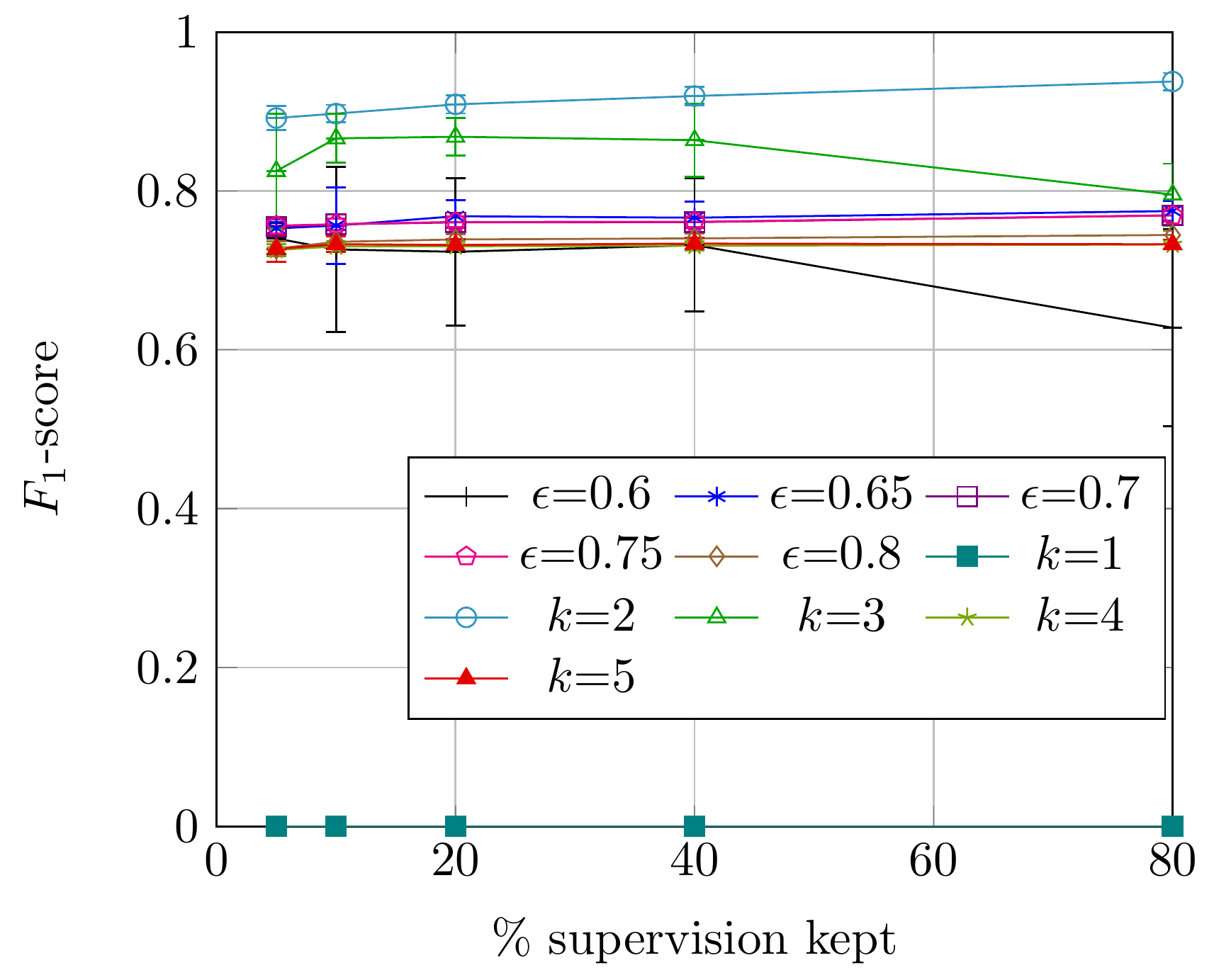}
\includegraphics[width=0.49\textwidth]{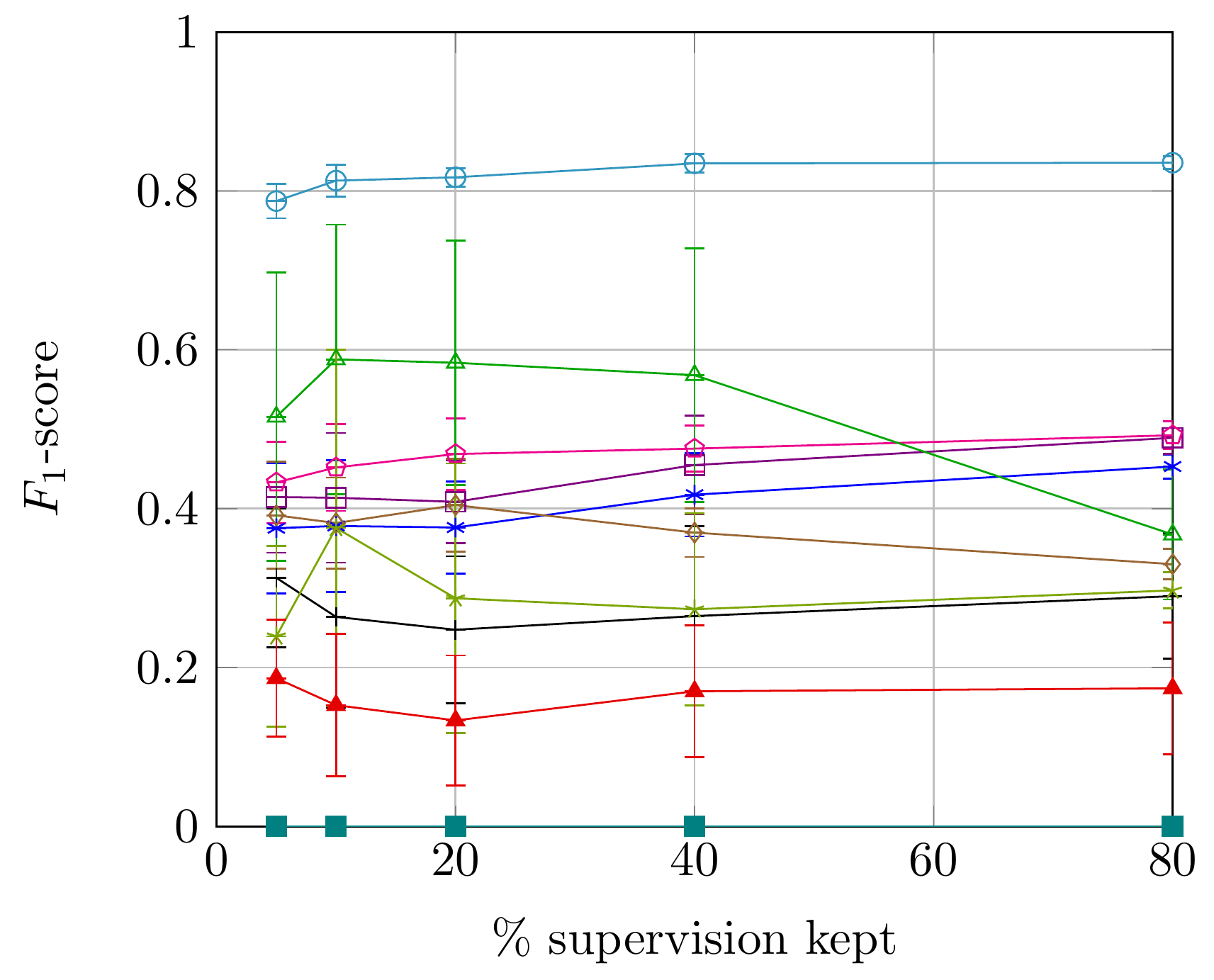}
\caption{$F_1$-score for \meet\ (left) and \move\ (right) as the supervision level increases per micro-batch.}
\label{fig:sc_labels_per_batch}
\end{figure}

Figure \ref{fig:sc_labels_per_batch} presents the results for all distinct values of $k$ and $\epsilon$ as the supervision level increases per micro-batch. The $F_1$-score is measured on the same test set for all supervision levels, namely the $20\%$ that remains unlabelled in the $80\%$ setting. The results indicate that $k{=}2$ is the best choice for both \meet\ and \move, achieving the highest accuracy on all supervision levels, including the low ones.

\begin{figure}[h]
\centering
\includegraphics[width=0.49\textwidth]{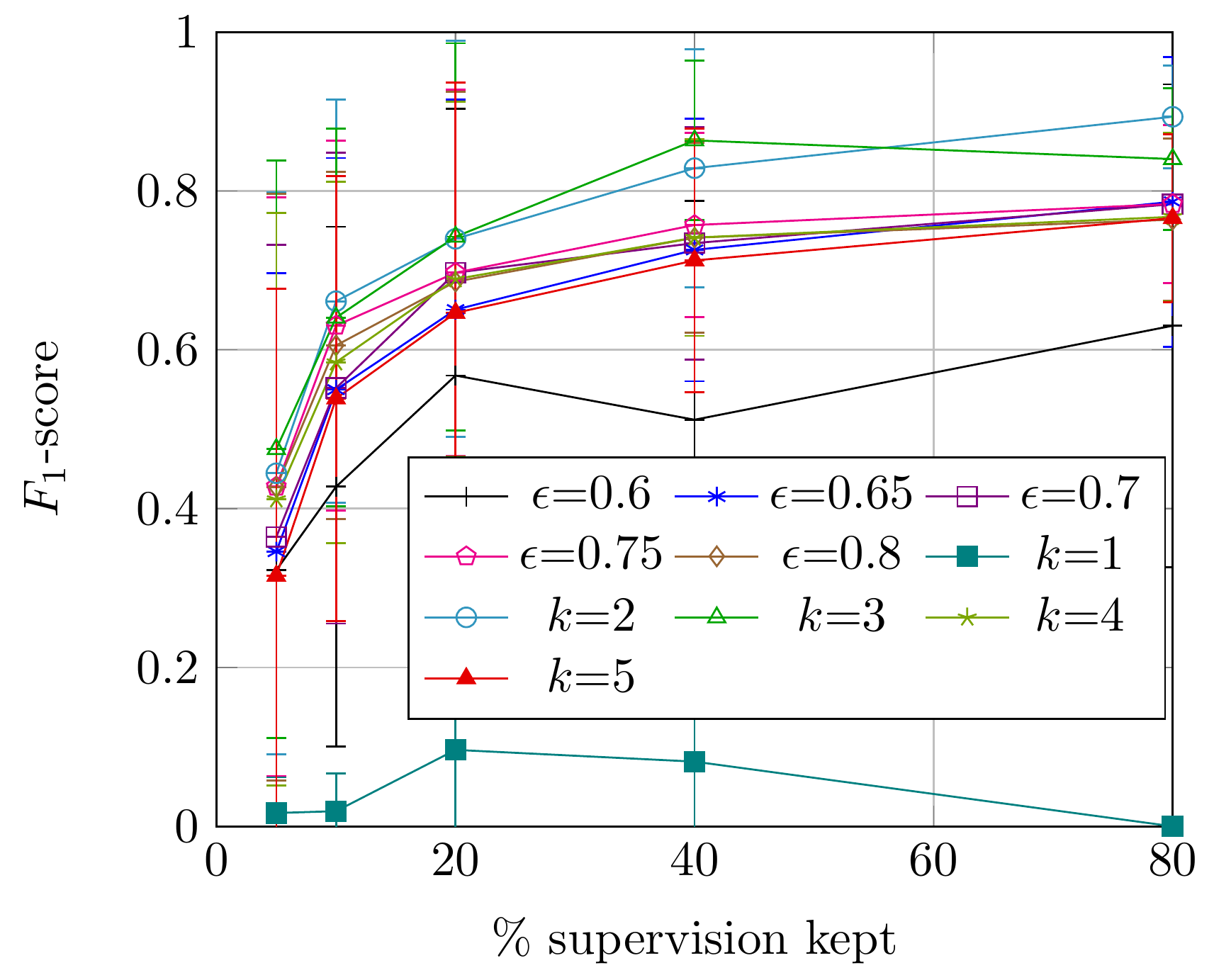}
\includegraphics[width=0.49\textwidth]{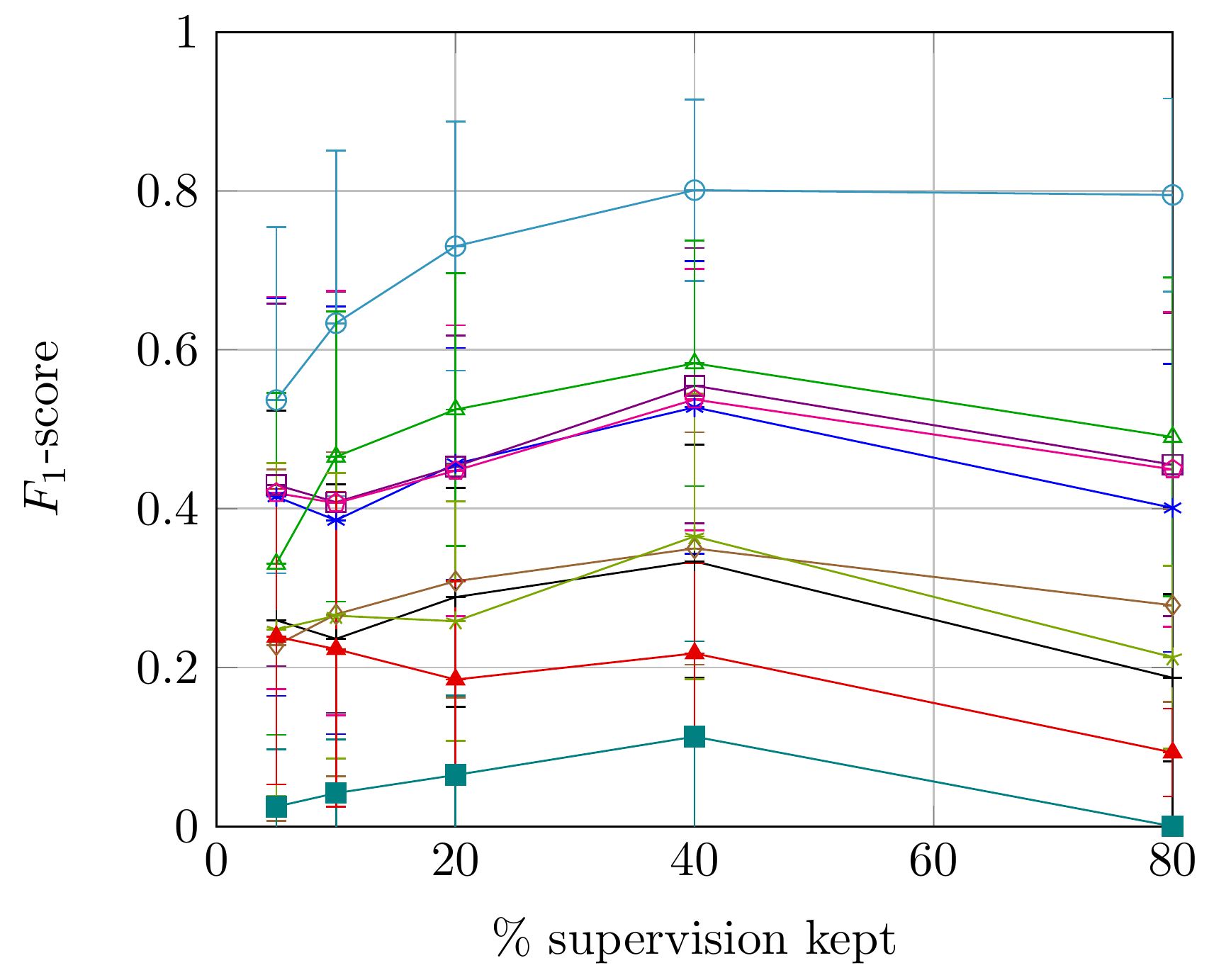}
\caption{$F_1$-score for \meet\ (left) and \move\ (right) as the level of supervision increases.}
\label{fig:sc_labels_over_batches}
\end{figure}

However, in a typical semi-supervised learning task, the assumption that every micro-batch contains some labels is too optimistic. A more realistic scenario is that a number of batches are completely labelled and the rest are completely unlabelled. We repeated the experiments, selecting uniformly a set of completely labelled batches, and present in Figure \ref{fig:sc_labels_over_batches} the results for different values of $k$ and $\epsilon$ as the supervision level increases.

Note that again $k{=}2$ is the best choice. As expected the $F_1$-score is lower in this more realistic setting, particularly for low supervision levels (e.g. $5\%$). However, the caching mechanism enables SPLICE to maintain a good accuracy despite the presence of completely unlabelled micro-batches. Another notable difference between Figure \ref{fig:sc_labels_per_batch} and Figure \ref{fig:sc_labels_over_batches} is that the standard error is now larger in most settings. This is because SPLICE is affected by the order in which labels arrive. It is also the reason why the standard error reduces as the supervision level increases. Based on these results, we chose $k{=}2$ for the rest of the evaluation.

\subsection{Experimental Results}

\subsubsection{Activity Recognition} \label{sec:caviar_eval}

\begin{figure}[t]
\centering
\includegraphics[width=0.49\textwidth]{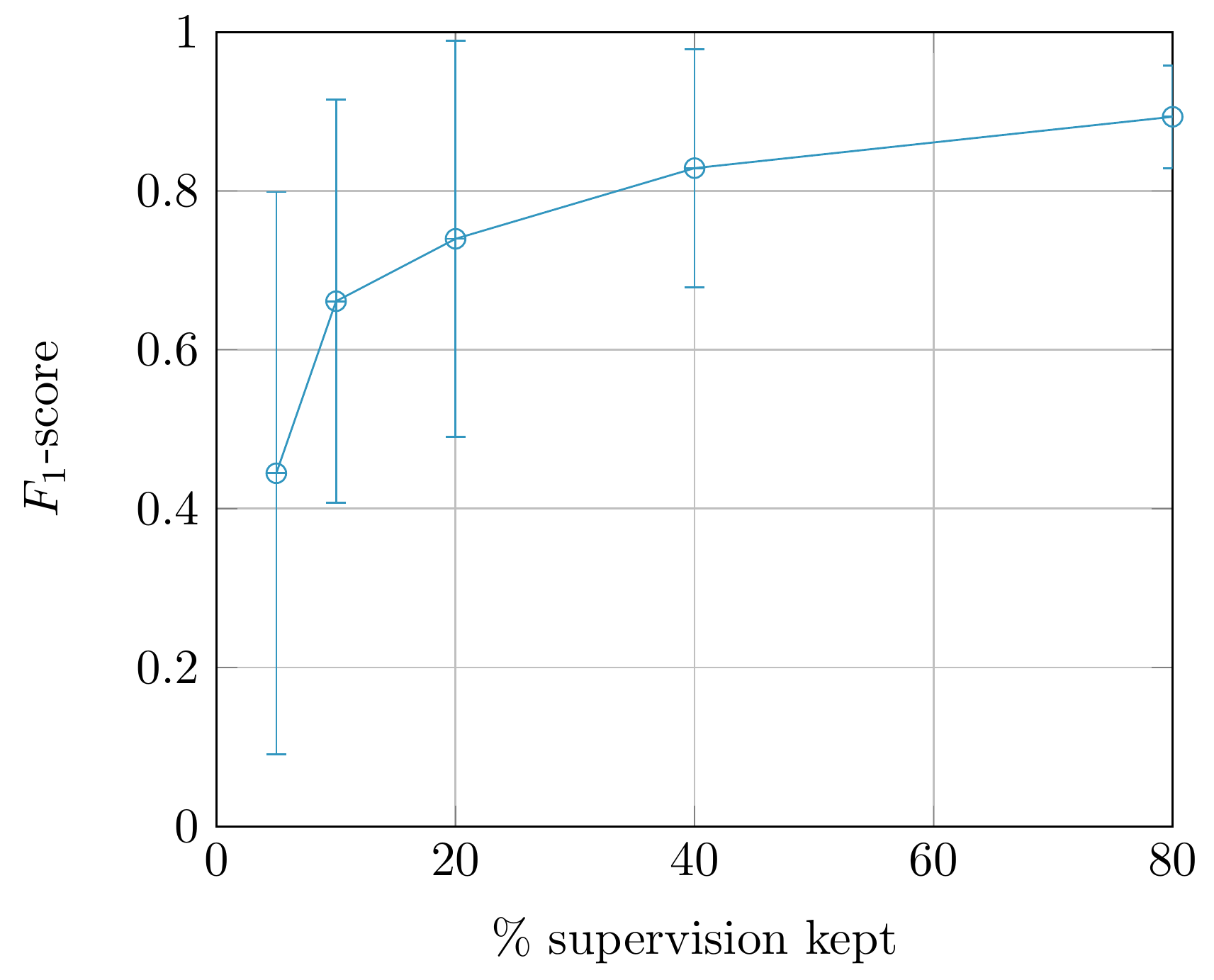}
\includegraphics[width=0.49\textwidth]{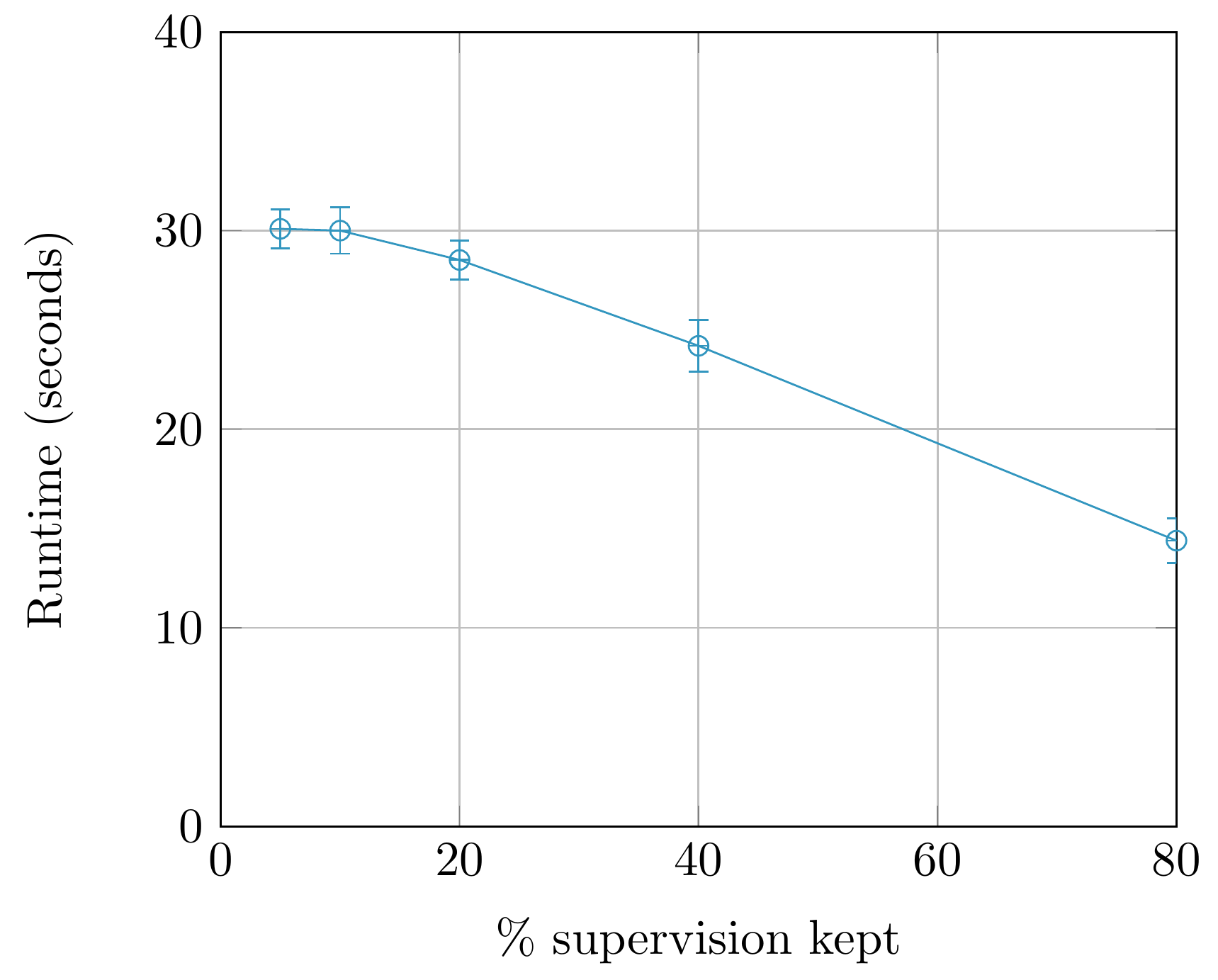}
\\
\includegraphics[width=0.49\textwidth]{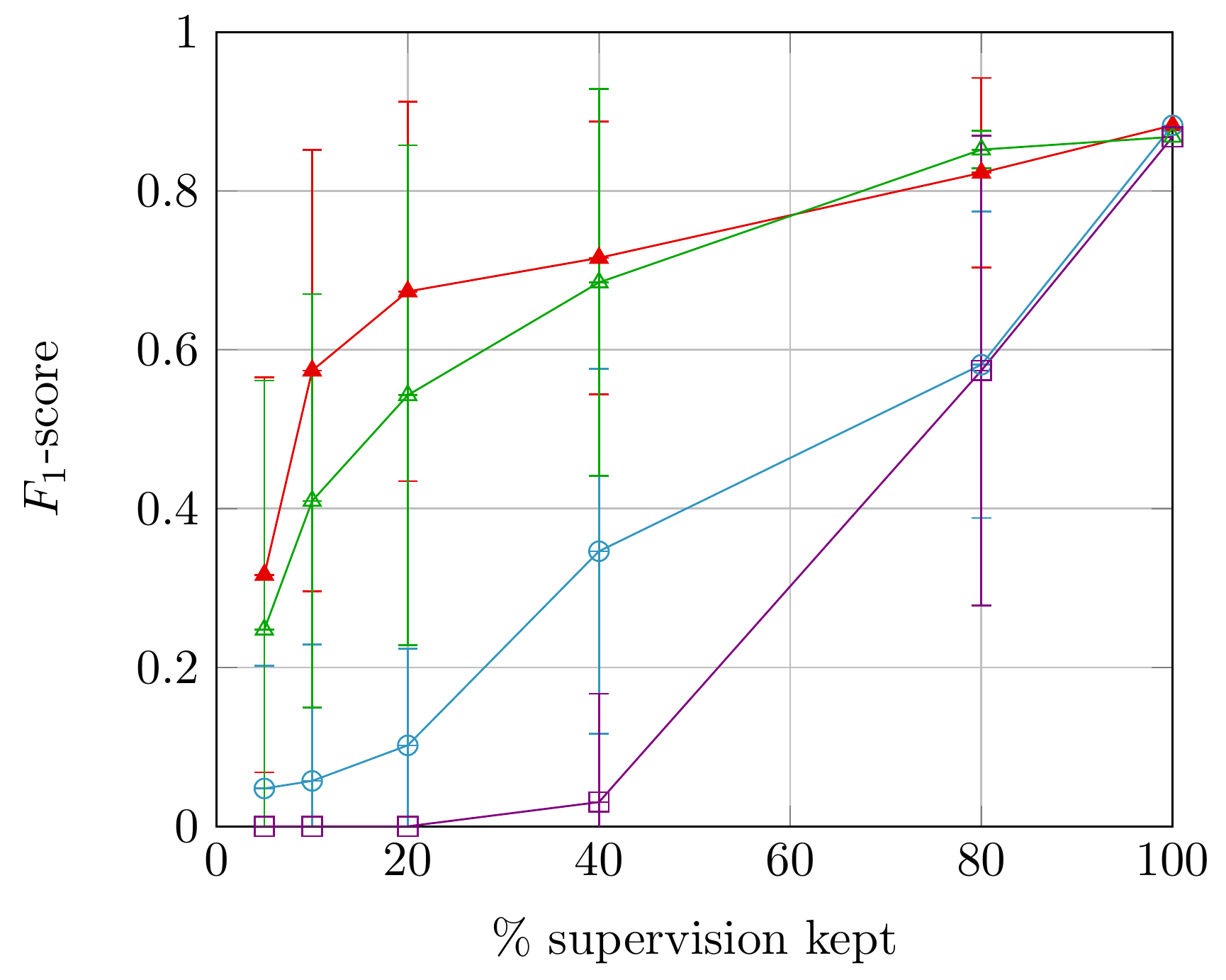}
\includegraphics[width=0.49\textwidth]{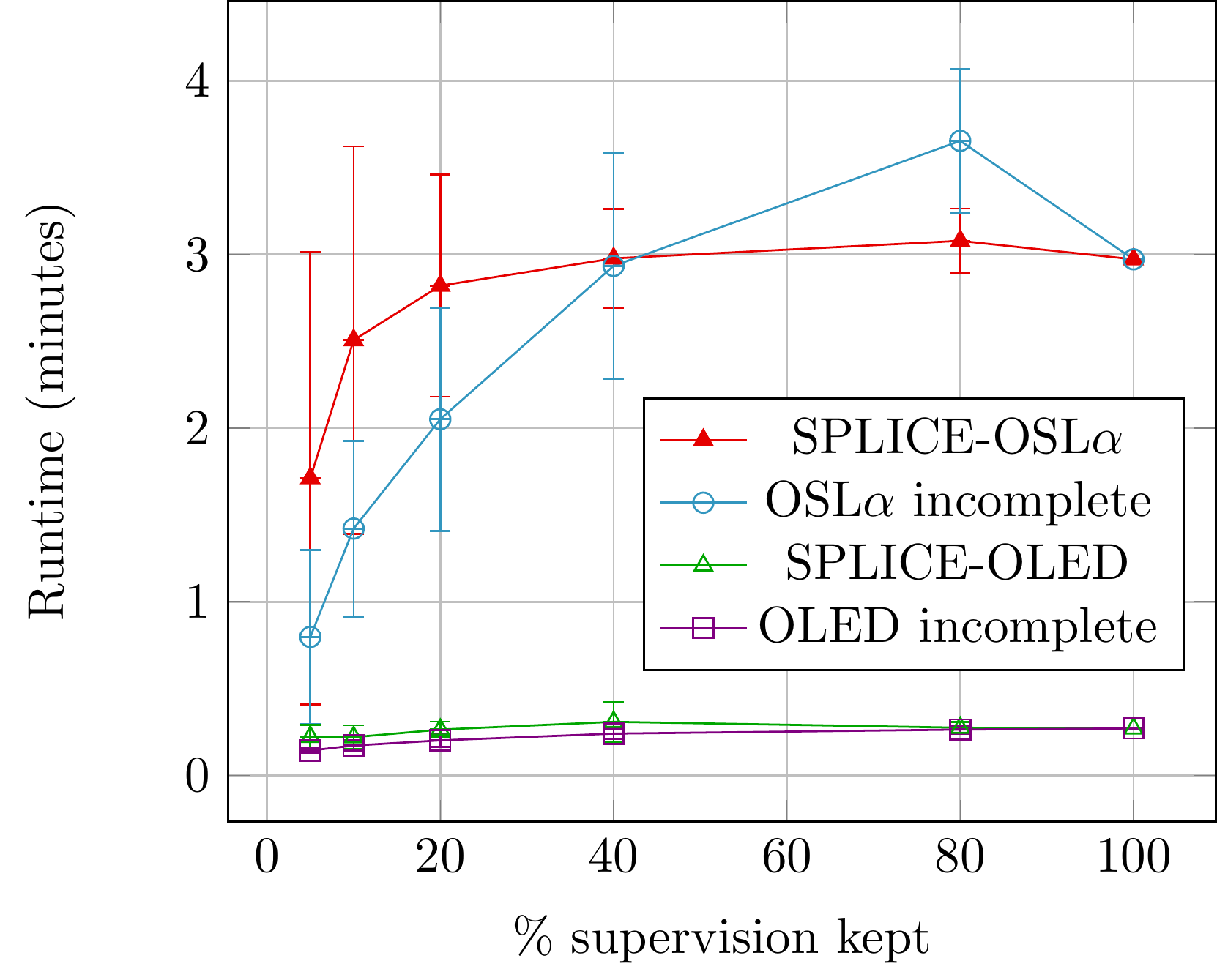}
\caption{$F_1$-score (left) and runtime (right) for \meet\ as the supervision level increases. Supervision completion (top) and semi-supervised online structure learning using both \OSLa\ and OLED (bottom).}
\label{fig:5}
\end{figure}

First, we tested the performance of SPLICE on the CAVIAR activity recognition dataset for both \meet\ and \move\ CEs, using the less optimistic scenario that a number of batches are completely labelled and the rest remain completely unlabelled. As in the hyperparameter selection process, the labelled micro-batches were selected using uniform sampling, while $20$ samples were taken at each supervision level. The results for \meet\ are presented in Figure \ref{fig:5}. The top figures present the $F_1$-score and runtime for the supervision completion, without structure learning, i.e., how well and how fast the true labels are recovered. The runtime of supervision completion is the total time required for completing all missing labels in each supervision level. To compute the $F_1$-score, however, only the $20\%$ that remains unlabelled in the $80\%$ supervision level is used. The bottom figures present the $F_1$-score and runtime (average training time per fold) of structure learning using \OSLa\ and OLED, i.e., how well and how fast the patterns for \meet\ and \move\ are learned. The $100\%$ setting in the bottom figures corresponds to full supervision, i.e., no unlabelled instances to be completed by SPLICE. In the bottom figures we also compare the performance of structure learning on the completed datasets against the datasets that contain unlabelled instances (incomplete).

\begin{figure}[t]
\centering
\includegraphics[width=0.49\textwidth]{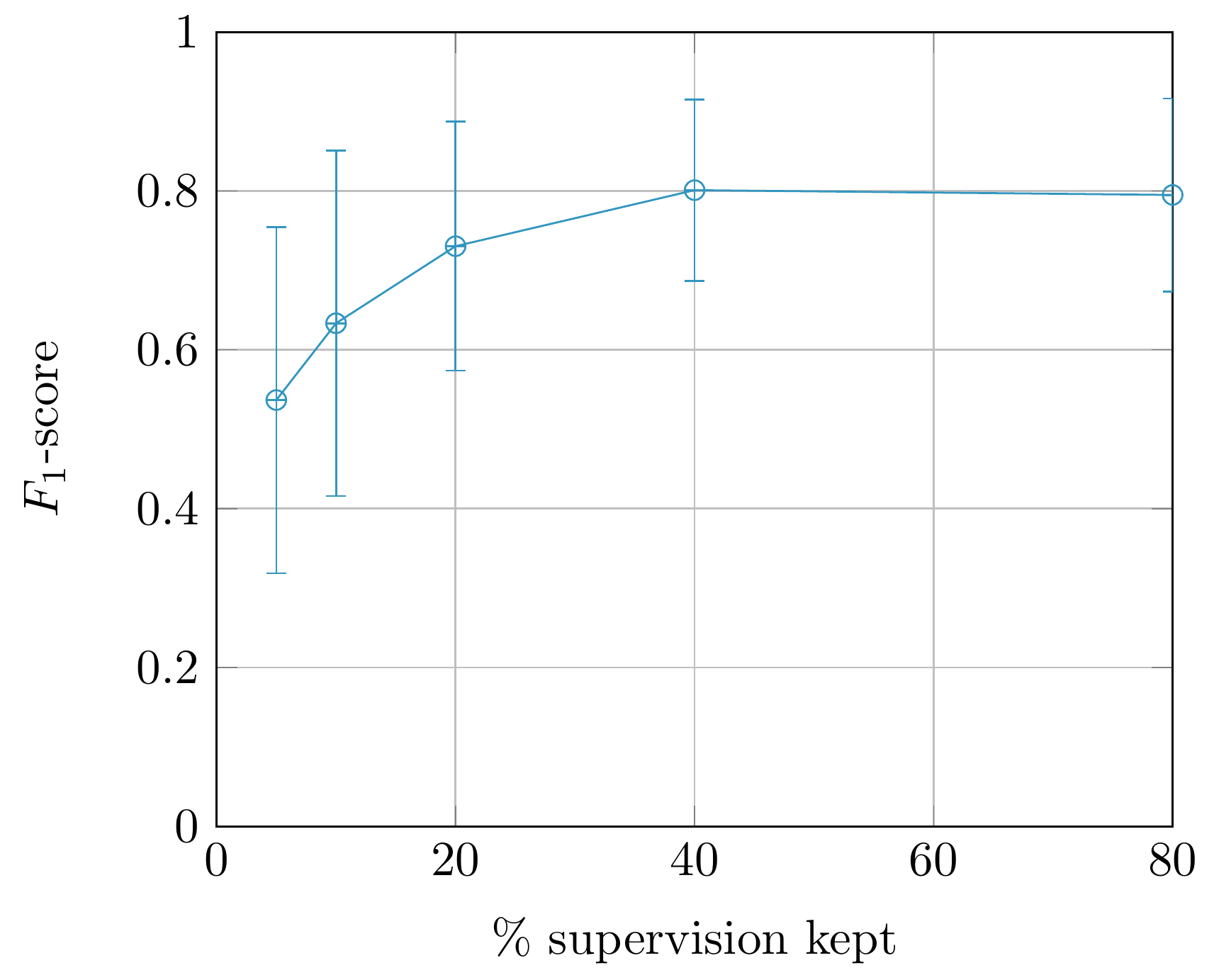}
\includegraphics[width=0.49\textwidth]{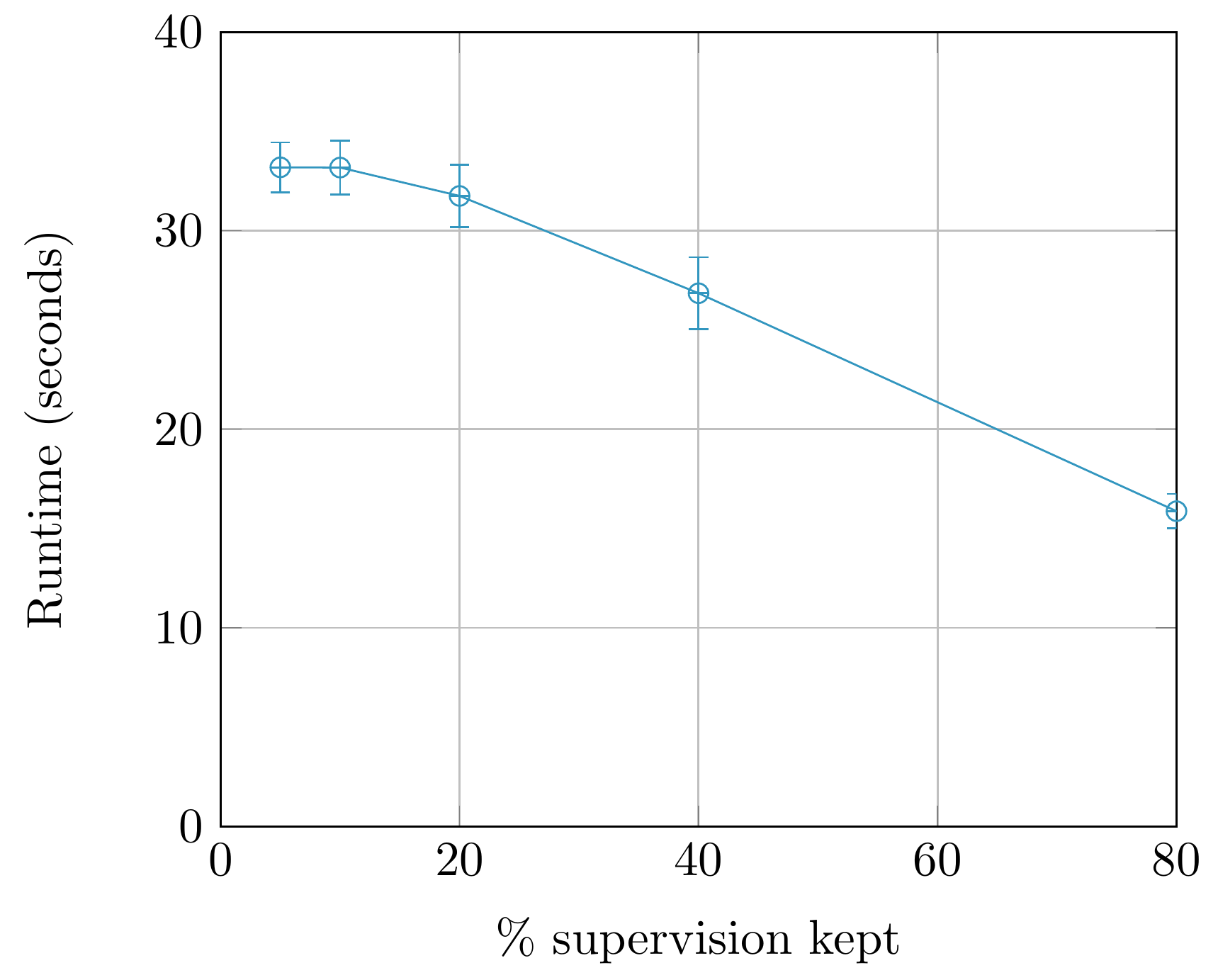}
\\
\includegraphics[width=0.49\textwidth]{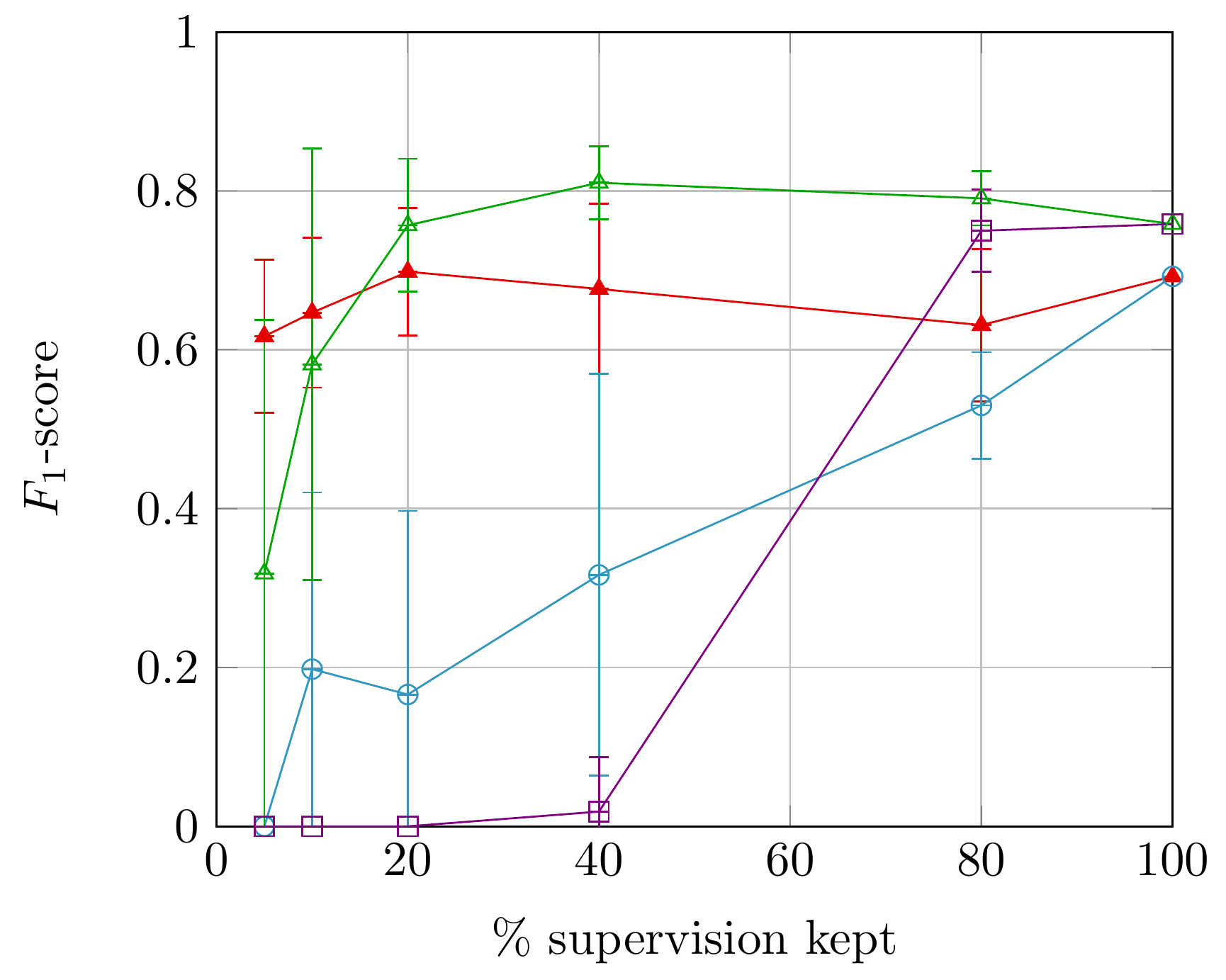}
\includegraphics[width=0.49\textwidth]{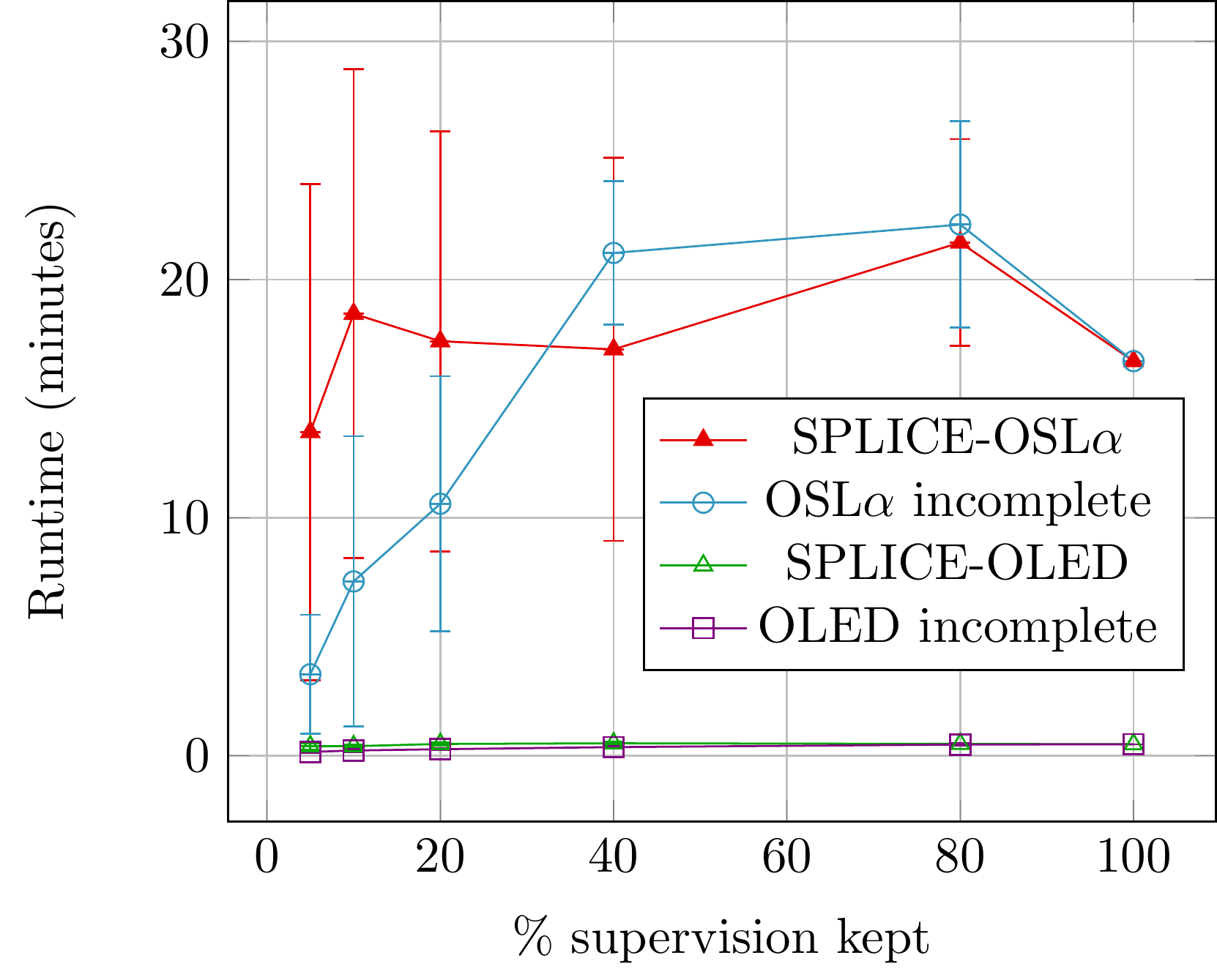}
\caption{$F_1$-score (left) and runtime (right) for \move\ as the supervision level increases. Supervision completion (top) and semi-supervised online structure learning using both \OSLa\ and OLED (bottom).}
\label{fig:6}
\end{figure}

Similar to the results shown in Section \ref{sec:selection}, we observe that  supervision completion effectively completes missing labels, even at low supervision levels. Also, the statistical error is reduced as the supervision level increases. The supervision completion runtime reduces as the supervision level increases, due to the smaller number of unlabelled instances that SPLICE needs to process. The results also suggest that SPLICE enhances substantially the accuracy of structure learning. The accuracy of both \OSLa\ and OLED without supervision completion is poor, due to the fact that the learners need to assume a label for the unlabelled instances, which is the negative label under the closed-world assumption. \OSLa\ achieves somewhat higher accuracy than OLED, especially for little given supervision, due to its ability to better handle noisy data. On the other hand, OLED is much faster than \OSLa.

Figure \ref{fig:6} presents the results for the \move\ CE, which mostly lead to the same conclusions as for the \meet\ CE, that is, that we can effective complete missing labels and consequently enhance significantly the accuracy of structure learning. One  difference in this setting is that \OSLa\ achieves higher accuracy than OLED only for low levels of supervision. Based on the results for both CEs, therefore, the version of SPLICE using \OSLa\ seems to be preferable for low supervision levels, but OLED has the advantage of being computationally more efficient.

\subsubsection{Maritime Monitoring} \label{sec:maritime_eval}

For the maritime monitoring dataset, we ran SPLICE using OLED because it provides better runtime performance than \OSLa\ on larger datasets. Recall that the maritime dataset comprises of $16{,}152{,}631$ SDEs, that is, approximately $1.2$GiB. Similar to the activity recognition evaluation, we used the less optimistic scenario, that assumes some micro-batches are completely labelled and the remaining ones are completely unlabelled. Due to the higher execution times, we performed experiments using only $5$ random splits of the data into labelled and unlabelled batches.

\begin{figure}[t]
\centering
\includegraphics[width=0.49\textwidth]{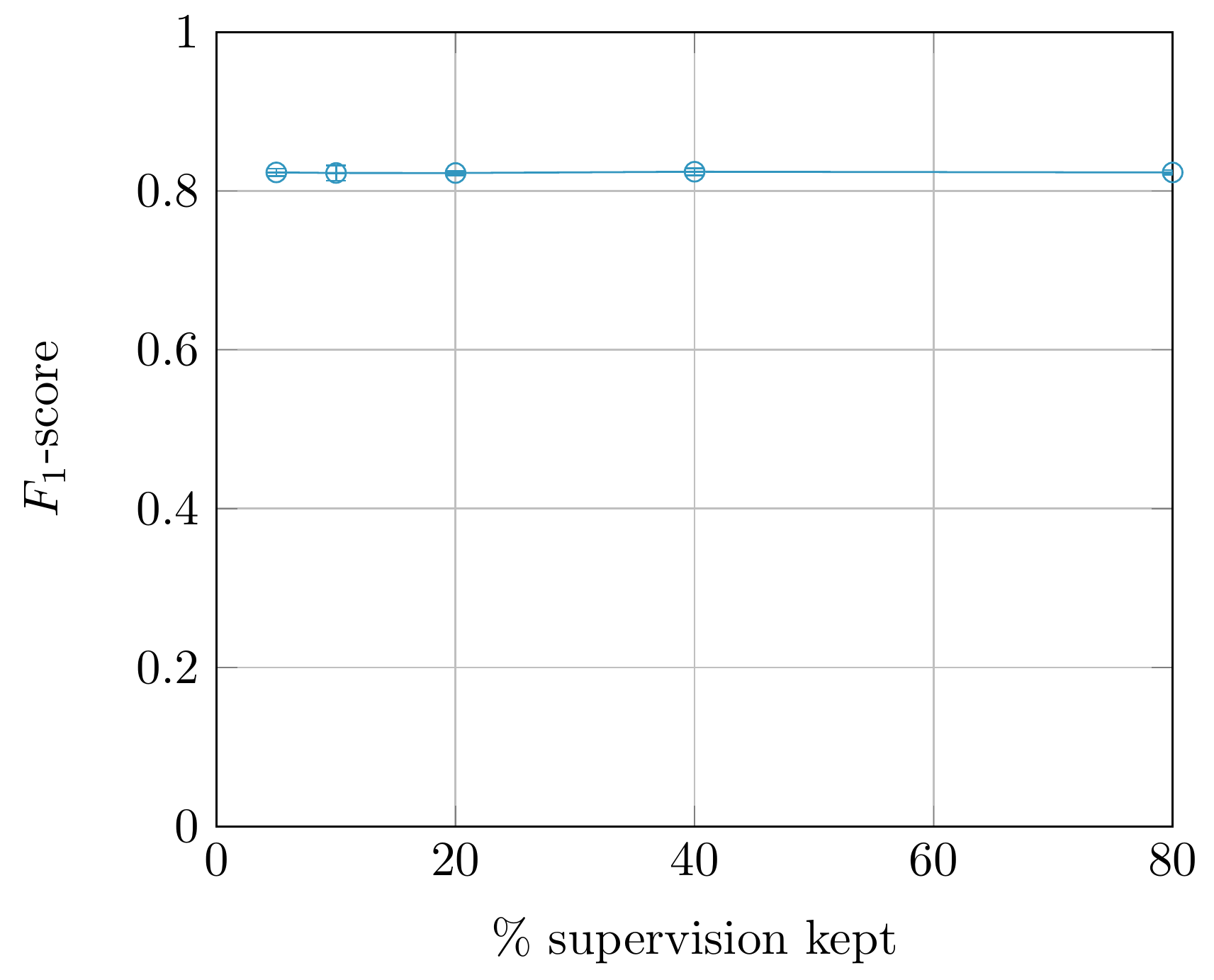}
\includegraphics[width=0.49\textwidth]{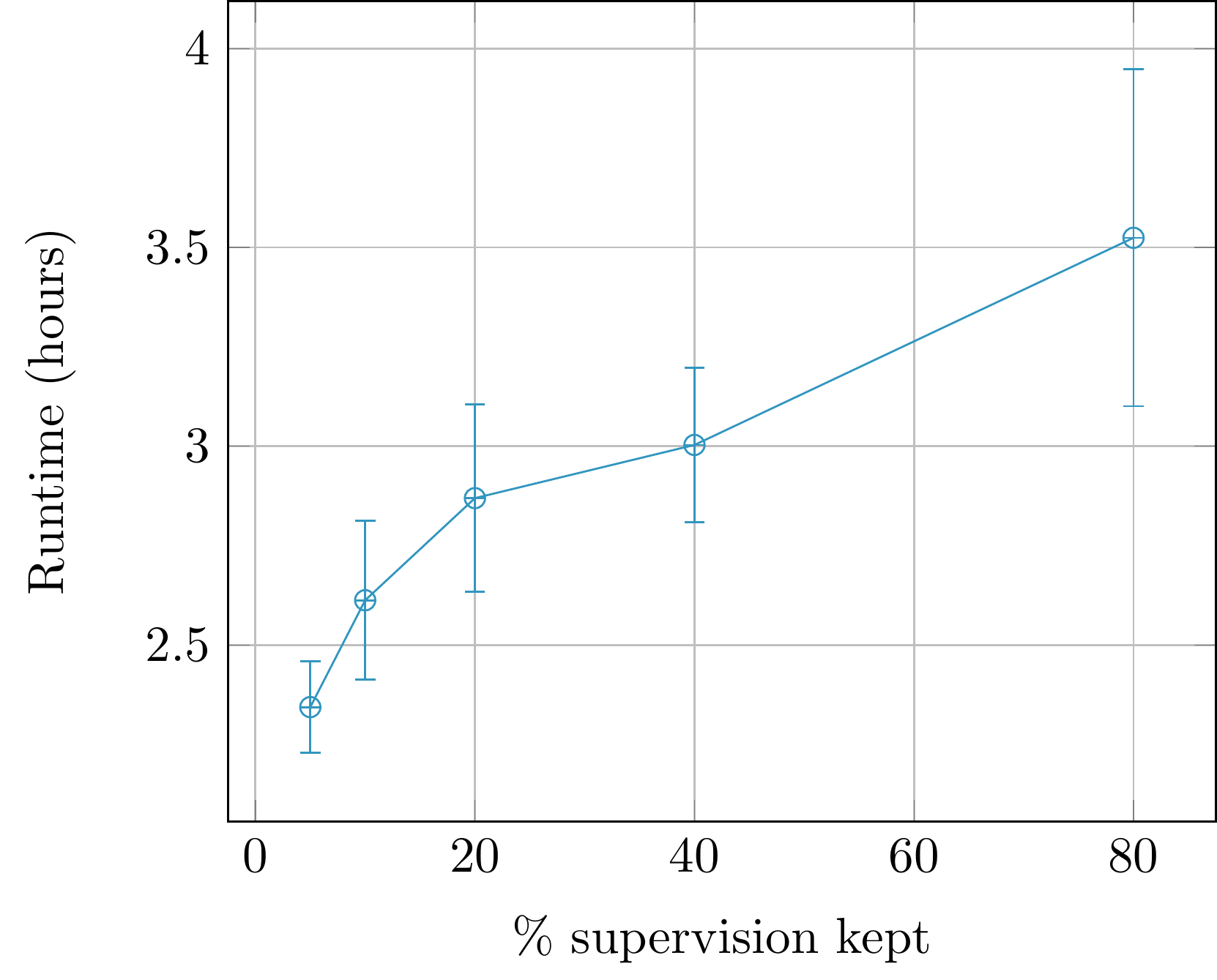}
\\
\includegraphics[width=0.49\textwidth]{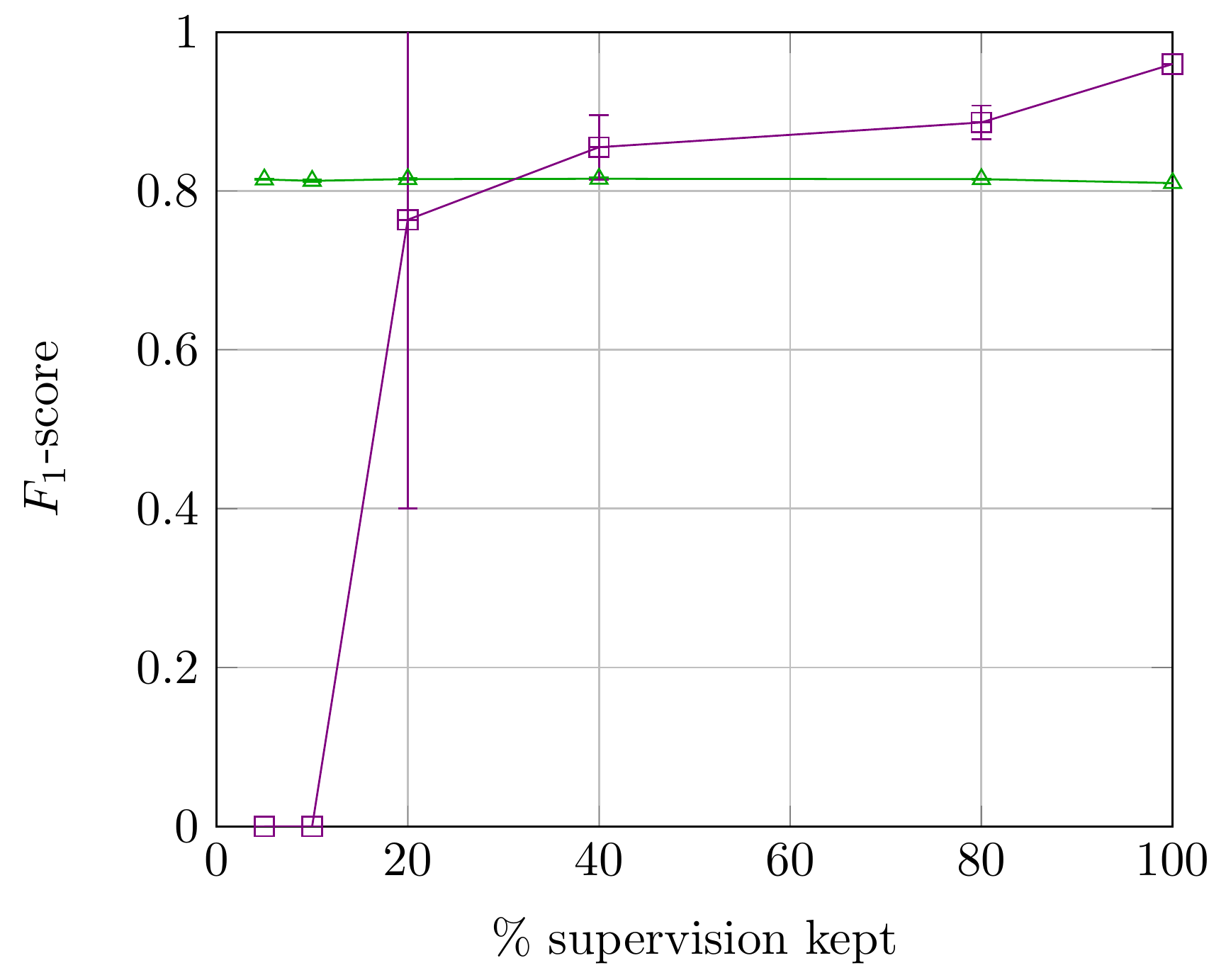}
\includegraphics[width=0.49\textwidth]{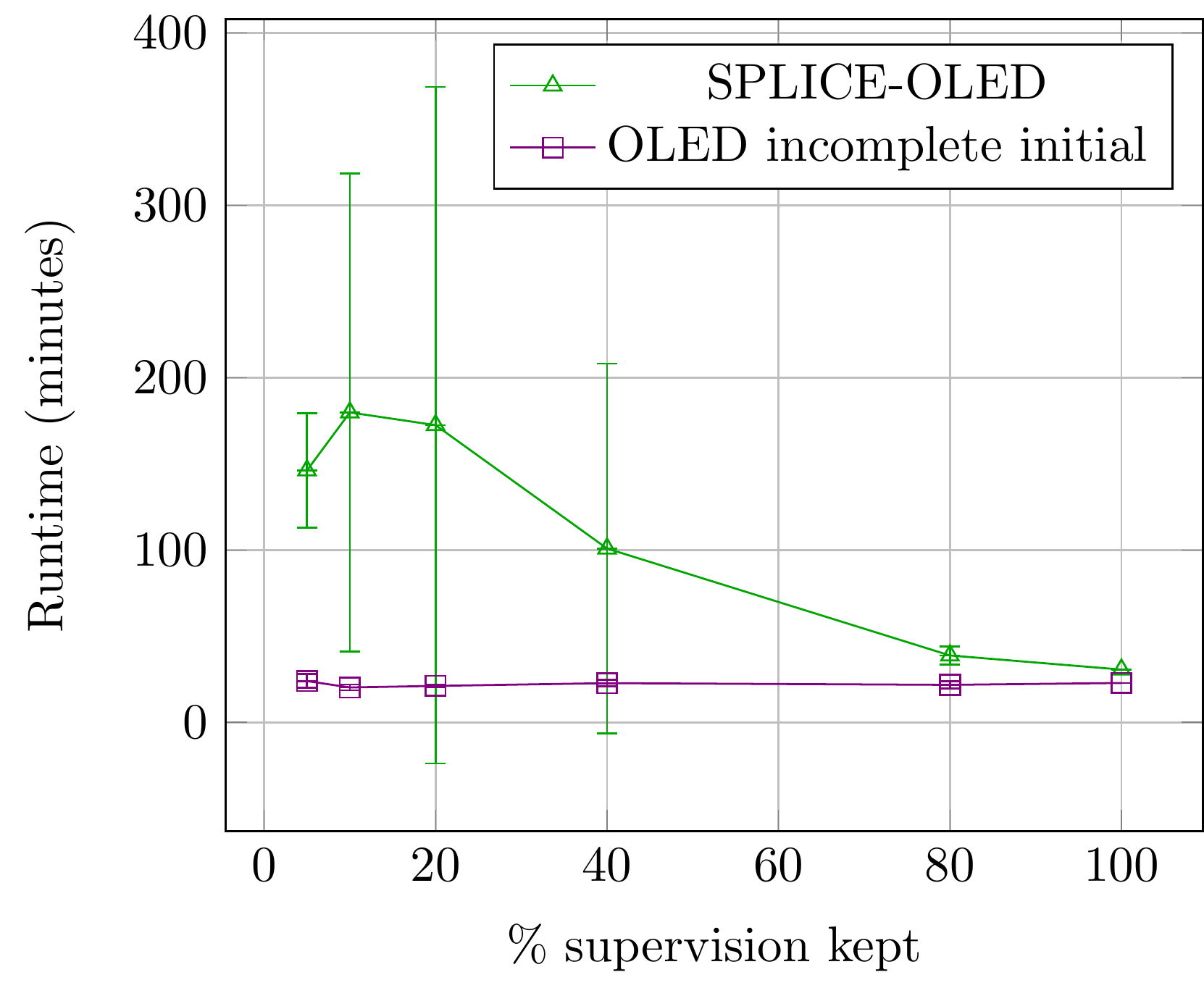}
\caption{$F_1$-score (left) and runtime (right) for \rendezvous\ as the supervision level increases. Supervision completion (top) and structure learning (bottom).}
\label{fig:7}
\end{figure}

The results for the \rendezvous\ CE are presented in Figure \ref{fig:7} in a similar form to the previous figures. The first observation is that the $F_1$-score of supervision completion (top left) is high even for $5\%$ of the given labels. On the other hand, the accuracy does not seem to change as the supervision level increases, i.e., there seems to be a ceiling to the number of labelled instances that can be correctly labelled. That stems from the fact that several positive examples share many common features (SDEs) with many negative examples, and are thus considered very similar. These examples are always misclassified, regardless of the given supervision level. The top-right diagram of Figure \ref{fig:7} shows that the supervision completion runtime increases along the supervision level. This is because the unique labelled examples cached by SPLICE greatly increase as the supervision increases, which was not the case with the activity recognition task. As a result, the quadratic term over cached labelled examples of the label caching component (see Section \ref{sec:cache_filter}) starts downgrading the computational cost.

Figure \ref{fig:7} also compares SPLICE along OLED, with OLED alone i.e., without supervision completion. Unlike the activity recognition experiments, OLED without SPLICE has been instructed to use only the starting points of each \rendezvous\ interval for structure learning. \rendezvous\ is very rarely re-initiated (approximately at $2\%$ of the time-points of a \rendezvous\ interval), and thus focusing on the starting points of the intervals when learning the initiating conditions of this concept can be helpful. Similarly, in the activity recognition experiments, OLED was instructed to use all data points, because \meet\ and \move\ are very frequently re-initiated (at $83\%$ and $65\%$ of the time-points of the intervals of these concepts), and thus it was highly desirable to use all available data for structure learning.

SPLICE with OLED operates as in the activity recognition experiments, that is, SPLICE labels all unlabelled examples, and then OLED uses all examples for structure learning. Instructing OLED to use only the starting points of the intervals does not improve performance in this case, since SPLICE makes some mistakes when labelling these points, compromising the performance of OLED. 

The bottom-left diagram of Figure \ref{fig:7} shows that SPLICE enhances considerably the accuracy of OLED in the common case of little supervision (below $20\%$). This is a notable result. In the case of $20\%$ supervision, OLED without SPLICE has a large deviation in performance, indicating the sensitivity of OLED in the presence of unlabelled data. On the other hand, SPLICE-OLED is very robust, as the standard deviation suggests. Provided with $40\%$ or more supervision, OLED without SPLICE can achieve better results. In these supervision levels and in the presence of very few re-initiations, it is better to proceed directly to structure learning, considering all unlabelled examples as negative.

With respect to efficiency, SPLICE-OLED seems to be much slower for lower supervision levels (see the bottom-right diagram of Figure \ref{fig:7}). This is expected because SPLICE-OLED uses many more examples than OLED alone in the maritime experiments.

\section{Related Work} \label{sec:related_work}


Structure learning is a task that has received much attention in the literature. The main approaches to this task stem either from probabilistic graphical models \citep{Pietra97inducingfeatures,Heckerman1999,McCallumCRF}, or Inductive Logic Programming (ILP) \citep{Quinlan90learninglogical,muggleton95,DeRaedt1997Claudien,BlockeelR98}. Online versions of structure learning methods have also been proposed, such as e.g., \citep{huynh2011osl,vagmcs2016osla,KatzourisAP16}, and some of them have been applied to real-world tasks, \citep{MichelioudakisA16,ArtikisKCBMSFP17}. All the aforementioned approaches, however, assume fully labelled training input in order to achieve generalisation. 


On the other hand, existing semi-supervised learning techniques \citep{ZhuSemiSupervised} attempt to exploit additional information provided by unlabelled data to guide the learning process, and enhance both performance and accuracy. These algorithms assume that training data are represented as propositional feature vectors. As a result, they cannot be directly applied to logic-based formalisms, that assume a relational data representation. Beyond expressiveness, typical approaches to semi-supervised learning also suffer from computational issues. For instance, self-training techniques \citep{Yarowsky95,GhahramaniJ93,Culp08,AlbinatiOOP15}, usually require a significant number of iterations over the training data to converge and thus are not appropriate for online learning. Co-training algorithms \citep{BlumM98,GoldmanZ00,Chawla05,ZhouL05} on the other hand, require that the training data are separated into distinct views, namely disjoint feature sets that provide complementary, ideally conditionally independent information about each instance, while each view alone is sufficient to accurately predict each class. Such limitations render many of these semi-supervised approaches incapable of handling the complexity of the relational structure learning task and inappropriate for online processing, which assumes a single pass over the training sequence.


Our proposed method is based on graph-based semi-supervised learning \citep{BlumMincut,ZhuGL03,BlumLRR04}, using a distance function that is suitable for first-order logic. A substantial amount of work exists in the literature on distance-based methods for learning from relational data. These approaches originate from  instance-based learning (IBL) \citep{AhaKA91}, which assumes that similar instances belong to similar classes (e.g. $k$NN). RIBL \citep{EmdeW96} extended IBL to the relational case by using a modified version of a similarity measure for logical atoms proposed by \cite{Bisson92}, together with a $k$NN classifier. \cite{Bisson92CC,Bisson92} uses a similarity measure, based on the structural comparison of logical atoms, to perform conceptual clustering. Although these distance measures have been used with success in several applications \citep{Bisson92,KirstenW98,KirstenW00}, they are limited to function-free Horn logic operating only over constants. Therefore, they require flattening of representations having non-constant terms, and thus cannot be easily applied to nested representations, such as the Event Calculus. \cite{BohnebeckHW98} improved RIBL to allow lists and other terms in the input representation, but their approach is not sensitive to the depth of the structure, i.e., functions.


Closest to SPLICE are techniques proposed for semi-supervised Inductive Logic Programming (ILP) and applied to web page classification. ICT (Iterative Cross-Training) \citep{SoonthornphisajK03} is a semi-supervised learning method, based on the idea of co-training. ICT uses a pair of learners, a strong and a weak one, to iteratively train each other from semi-supervised training data. Each learner receives an amount of labelled and unlabelled data. The strong learner starts the learning process from the labelled data, given some prior knowledge about the domain, and classifies the unlabelled data of the weak learner. The weak learner, which has no domain knowledge, then uses these recently labelled data produced by the strong learner, to learn and classify the unlabelled data of the strong learner. This training process is repeated iteratively. ICT-ILP \citep{SoonthornphisajK04} is an extension of ICT that uses an ILP system as one of the classifiers, that is, the strong learner, and a Naive Bayes classifier for the weak learner. The ILP system makes use of a background knowledge that encodes the prior domain knowledge and induces a set of rules from the labelled data. These rules are used to classify the unlabelled examples of the weak learner. \cite{LiG11,LiG12} proposed a similar approach based on relational tri-training. Three different relational learning systems, namely Aleph \citep{SrinivasanAleph}, kFOIL \citep{LandwehrPRF06} and nFOIL\citep{LandwehrKR07}, are initialised using the labelled data and background knowledge. Then the three classifiers are refined by iterating over the unlabelled data. At each iteration, each unlabelled example is labelled by the three classifiers. In case two of them agree on the labelling of the example, then this example is labelled accordingly. The final classification hypothesis is produced via majority voting of the three base classifiers.

The aforementioned approaches to semi-supervised structure learning iterate multiple times over the training data in order to generalise. Therefore, they are not suitable for \emph{online} structure learning. Consequently, the method presented in this paper is the first to tackle the problem of online semi-supervised structure learning.

\section{Conclusions and Future Work} \label{sec:conclusions}

We presented SPLICE, a novel approach to online structure learning that operates on partially supervised training sequences. SPLICE completes the missing supervision continuously as the data arrive in micro-batches, and can be combined with any online supervised structure learning system. As it processes the input stream, SPLICE can cache previously seen labelled examples for future usage and filter noisy, contradicting labelled examples that may compromise the overall accuracy of the structure learning task. Experimental results in the domain of composite event recognition, using a benchmark dataset for activity recognition and a real dataset for maritime monitoring, showed that SPLICE can enable the underlying structure learner to learn good models even in the presence of little given annotation.

We are currently investigating various extensions to SPLICE, including the improvement of the distance function, especially in the case of many unrelated features. Moreover, we are examining the possibility of extending SPLICE with abductive inference, in order to perform structure learning on hidden concepts with partially supervised data. This last extension is desirable for learning CE definitions, because, usually, the provided labels are different from the target concept.

\begin{acknowledgements}
The work has been funded by the EU H2020 project datAcron (687591). We would also like to thank Nikos Katzouris for providing assistance on the distance functions for first-order logic and helping us running OLED.
\end{acknowledgements}

\bibliographystyle{spbasic}
\bibliography{references}

\end{document}